\documentclass[10pt,twocolumn,letterpaper]{article}

\usepackage[pagenumbers]{cvpr} 

\usepackage[dvipsnames]{xcolor}

\usepackage{dsfont}
\usepackage{xcolor}
\usepackage{color, colortbl}
\usepackage{booktabs}
\usepackage{multirow}
\usepackage{makecell}
\usepackage{pifont}
\usepackage{amssymb}
\usepackage{subcaption}
\usepackage{amsmath}

\definecolor{color1}{HTML}{ECF4F9}
\definecolor{color2}{HTML}{FFF1E0}
\definecolor{color3}{HTML}{ECF4E9}

\definecolor{Gray}{gray}{0.9}
\definecolor{LightBlue}{RGB}{236,244,249}

\newcommand{\CC}[1]{\cellcolor{LightBlue}}
\newcommand{\RC}[1]{\rowcolor{LightBlue}}

\newcommand{\cmark}{\color{green}\ding{51}}%
\newcommand{\xmark}{\color{red}\ding{55}}%

\DeclareMathOperator*{\argmax}{arg\,max}

\usepackage{algorithm}
\usepackage{algpseudocode}
\algrenewcommand{\algorithmicrequire}{\textbf{Input:}}
\algrenewcommand{\algorithmicensure}{\textbf{Output:}}

\definecolor{cvprblue}{rgb}{0.21,0.49,0.74}
\usepackage[pagebackref,breaklinks,colorlinks,citecolor=cvprblue]{hyperref}

\def\paperID{3405} 
\def\confName{CVPR}
\def\confYear{2024}

\title{Active Generalized Category Discovery}

\author{Shijie Ma$^{1,2}$, 
	~Fei Zhu$^{3}$,
        ~Zhun Zhong$^{4,5}$,
        ~Xu-Yao Zhang$^{1,2}$\footnotemark[1]~,
        ~Cheng-Lin Liu$^{1,2}$\\
	$^1$MAIS, Institute of Automation, Chinese Academy of Sciences, China\\
	$^2$School of Artificial Intelligence, University of Chinese Academy of Sciences, China\\
        $^3$Centre for Artificial Intelligence and Robotics, HKISI-CAS, China \\
        $^{4}$School of Computer Science and Information Engineering, Hefei University of Technology, China \\
        $^{5}$School of Computer Science, University of Nottingham, NG8 1BB Nottingham, UK \\
	{\tt\small \{mashijie2021, zhufei2018\}@ia.ac.cn, zhunzhong007@gmail.com, \{xyz, liucl\}@nlpr.ia.ac.cn}
}

\begin{document}

\twocolumn[{%
\renewcommand\twocolumn[1][]{#1}%
\maketitle
\begin{center}
    \centering
    \captionsetup{type=figure}
    \vspace{-10pt}
    \includegraphics[width=.95\textwidth]{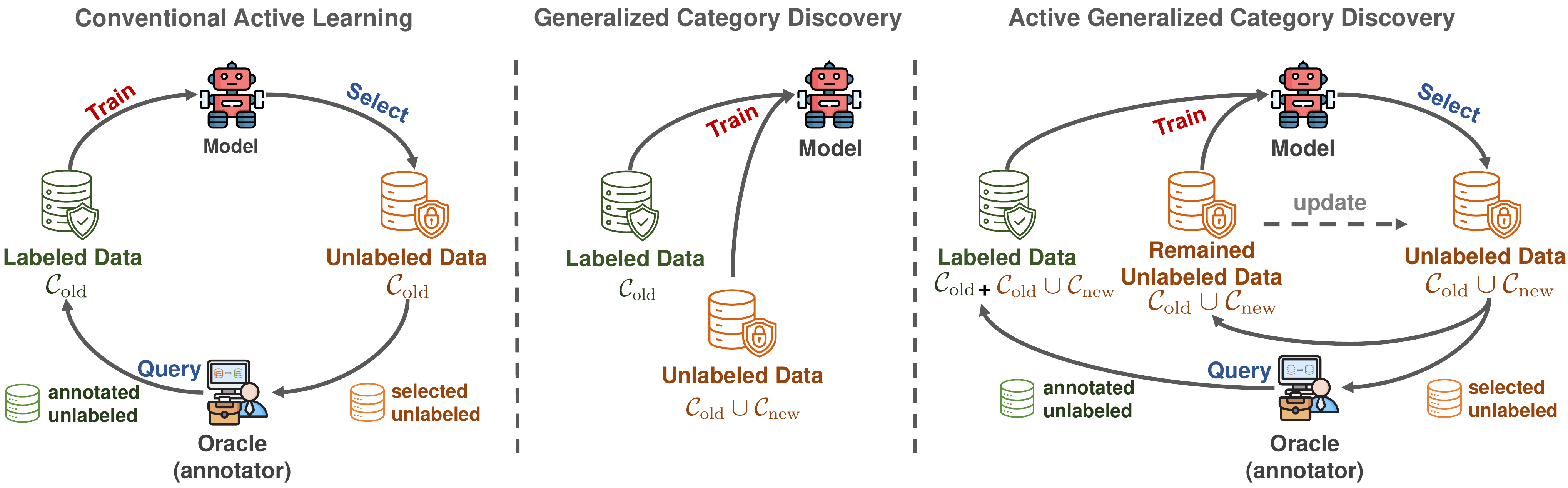}
    \vspace{-10pt}
    \captionof{figure}{The diagram of three settings. Left: Conventional AL is a closed-world setting, where labeled and unlabeled classes are identical. Middle: GCD requires no active labeling and suffers from severe issues. Right: AGCD is an open-world extrapolated version of AL, where unlabeled data contains novel categories, and models are trained on both labeled and unlabeled data to cluster both old and new classes.}
    \label{fig:agcd-setting}
\end{center}
}]

{\renewcommand{\thefootnote}{\fnsymbol{footnote}}
\footnotetext[1]{Corresponding author.}}


\begin{abstract}
\vspace{-5pt}
Generalized Category Discovery (GCD) is a pragmatic and challenging open-world task, which endeavors to cluster unlabeled samples from both novel and old classes, leveraging some labeled data of old classes. Given that knowledge learned from old classes is not fully transferable to new classes, and that novel categories are fully unlabeled, GCD inherently faces intractable problems, including imbalanced classification performance and inconsistent confidence between old and new classes, especially in the low-labeling regime. Hence, some annotations of new classes are deemed necessary. However, labeling new classes is extremely costly. To address this issue, we take the spirit of active learning and propose a new setting called Active Generalized Category Discovery (AGCD). The goal is to improve the performance of GCD by actively selecting a limited amount of valuable samples for labeling from the oracle. To solve this problem, we devise an adaptive sampling strategy, which jointly considers novelty, informativeness and diversity to adaptively select novel samples with proper uncertainty. However, owing to the varied orderings of label indices caused by the clustering of novel classes, the queried labels are not directly applicable to subsequent training. To overcome this issue, we further propose a stable label mapping algorithm that transforms ground truth labels to the label space of the classifier, thereby ensuring consistent training across different active selection stages. Our method achieves state-of-the-art performance on both generic and fine-grained datasets. Our code is available at \url{https://github.com/mashijie1028/ActiveGCD}
\end{abstract}    
\vspace{-.15in}
\section{Introduction}
\label{sec:introduction}

Humans could transfer previously acquired knowledge while learning new concepts~\cite{macaulay2002transfer}. For example, once children have been taught to recognize ``cats'' and ``dogs'' based on external contours, they can group ``birds'' and ``bears'' according to the same rule. However, due to species disparities, this classification criteria is limited. Children may confuse ``zebras'' with ``horses'', and ``huskies'' with ``wolves'', and they still need guidance to focus on fine-grained features like stripes and eyes. Thus, proper guidance is indispensable when acquiring new knowledge. However, seeking help every time is impractical. Instead, they are supposed to actively query some confusing samples~\cite{castro2008human}.

\begin{figure}[!t]
    \centering
    \includegraphics[width=.75\linewidth]{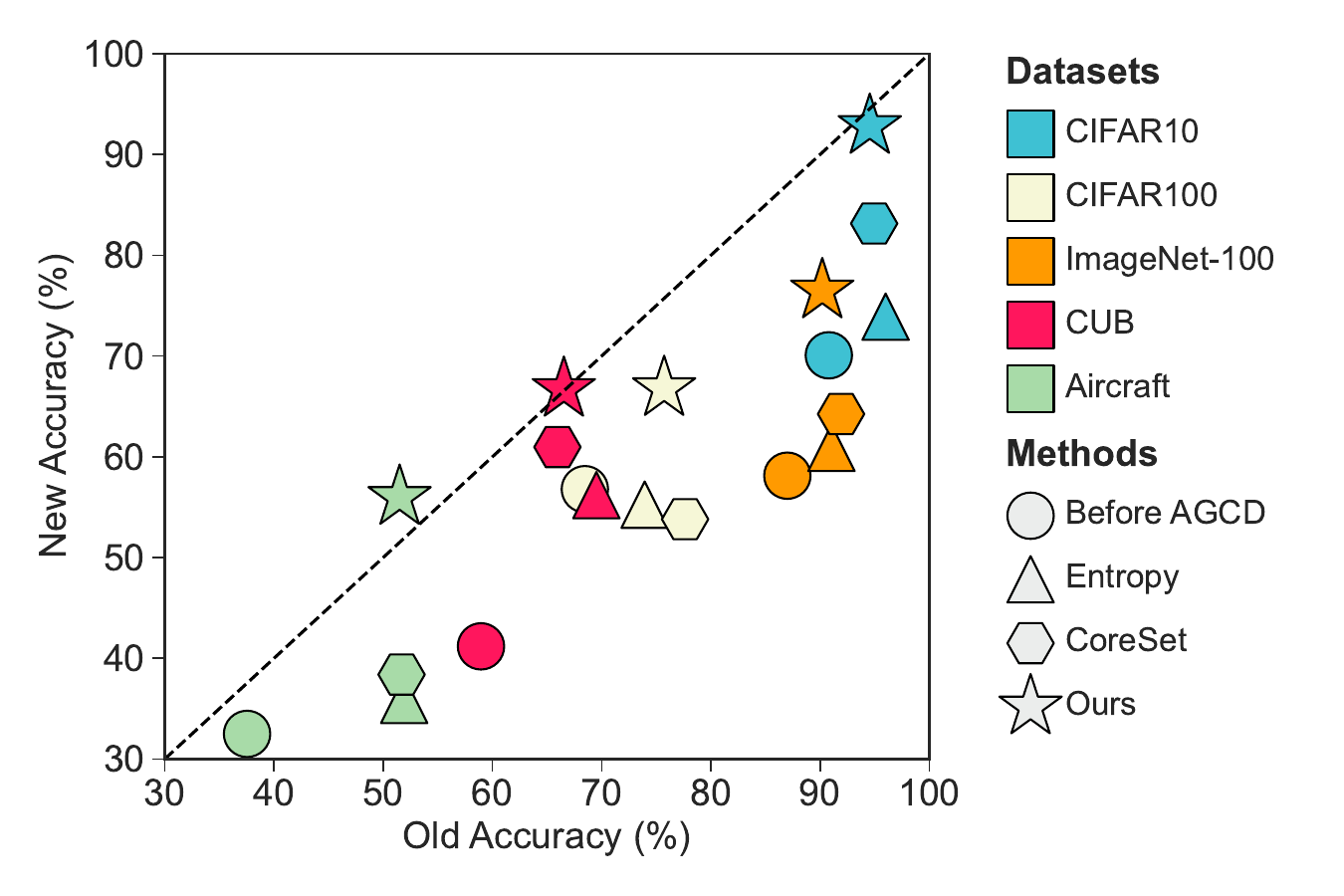}
    \vspace{-7pt}
    \caption{Accuracy of old and new classes in AGCD with different methods (shapes) on various datasets (colors). The closer to the diagonal, the more balanced accuracy between old and new classes. In each dataset (color), our method (star) achieves not only the best overall performance but also more balanced accuracy.}
    \vspace{-13pt}
    \label{fig:overall-old-new-acc}
\end{figure}

Deep learning is also inspired by the cognitive processes above, and could be endowed with the abilities of knowledge transfer~\cite{pan2009survey,han2019learning,zhuang2020comprehensive} and active learning~\cite{settles2009active,ren2021survey,zhan2022comparative}, especially in the open-environments~\cite{9040673,vaze2022openset,ma2023towards,yang2021generalized,zhu2024openworld} containing unlabeled novel categories. In this paper, we study the task of Generalized Category Discovery (GCD)~\cite{vaze2022generalized, pu2023dynamic, Wen_2023_ICCV, Zhao_2023_ICCV}, which aims to transfer knowledge from some labeled samples of old classes to cluster novel categories in the unlabeled data. In addition, models should also be able to classify old classes present in the unlabeled data.  Pioneer works~\cite{vaze2022generalized,fei2022xcon} leverage supervised~\cite{khosla2020supervised} and unsupervised contrastive learning~\cite{chen2020simple} with non-parametric K-Means~\cite{macqueen1967some} for clustering. Later works~\cite{pu2023dynamic,zhang2023promptcal} further exploit underlying cross-instance relationships. Wen \etal~\cite{Wen_2023_ICCV} rethink the failure of parametric classifiers and propose a simple method to achieve impressive results.

Although great progress has been made, GCD still faces intractable problems, including imbalanced accuracy (see Fig.~\ref{fig:old-new-acc-imbalance}) and inconsistent confidence (see Fig.~\ref{fig:old-new-confidence}) between old and new classes, especially in low-labeling regimes.
In essence, these issues arise from the nature of the GCD task itself. As old knowledge is not fully transferable to the new one, and novel classes are fully unlabeled, models would encounter inherent challenges, and could not rectify errors by themselves without the supervision of confusing categories. Therefore, we argue that some annotations of new categories~\cite{zhu2021prototype} are necessary. However, due to the computational cost of annotation, it is not practical to label all the novel classes. This raises a question: \emph{Can deep learning models actively select a small number of unlabeled samples for labeling to remarkably enhance category discovery?}

In this work, we try to answer this question and propose a new setting, namely Active Generalized Category Discovery (AGCD) as in Fig.~\ref{fig:agcd-setting}. During training, models actively select a limited number of samples in unlabeled data, which contains both old and new classes, and query their labels from the oracle, these newly-labeled data are then incorporated into labeled data for the next training round. Through human-in-the-loop interaction, models actively select informative novel samples, acquire knowledge that could not be obtained via pure unsupervised learning, and rectify previous errors and biases. AGCD is a realistic setting, which addresses the problems of GCD and largely enhances the performance, requiring very limited annotations. As in Fig.~\ref{fig:overall-old-new-acc}, we improve the new accuracy of GCD by 25.52\%/23.49\% on CUB/Air with only $\sim 2.5$ samples labeled per class.

In the task of AGCD, one could inevitably encounter two challenges, which we aim to address in this paper: (1) Conventional AL methods do not take novel categories into consideration, which makes them not applicable to AGCD and leads to sub-optimal results. (2) Considering the clustering nature of GCD, the queried ground truth labels could not be directly used by parametric classifiers due to the different ordering of indices. To solve the first problem, we take novelty, informativeness and diversity into consideration and propose an adaptive sampling strategy called \texttt{Adaptive-Novel}, which adaptively chooses samples within appropriate uncertainty intervals according to the clustering performance. To alleviate the second problem, we propose to perform label mapping on the queried samples which ``translates'' ground-truth labels to the labels the model could understand, however, considering the scarcity of labeled data, we devise a stable label mapping method with the model exponential moving average~\cite{tarvainen2017mean,zhu2022rethinking,NEURIPS2023_98143953}.

Our contributions are summarized as follows: (1) We propose a new task called Active Generalized Category Discovery (AGCD) considering the inherent issues in GCD, and establish its pipeline and metrics. (2) We propose an adaptive query strategy called \texttt{Adaptive-Novel} to select valuable novel samples for labeling and address the problems of GCD with affordable budgets. (3) We devise a stable label mapping method to obtain credible mapping and alleviate the issue of different label ordering in clustering. (4) Extensive experiments show our method achieves state-of-the-art performance among various strategies on generic and fine-grain datasets, as in Fig.~\ref{fig:overall-old-new-acc}.

\section{Related Works}
\label{sec:related-works}

\textbf{Novel Category Discovery} (NCD)~\cite{troisemaine2023novel} was first formalized as deep transfer clustering~\cite{han2019learning} to discover unlabeled new classes using the knowledge of labeled classes. Han \etal~\cite{han2019learning,Han2020Automatically} utilize self-supervision for representation learning and ranking statistics for knowledge transfer. Zhong \etal~\cite{zhong2021openmix} propose to mixup~\cite{zhang2018mixup} old and new classes to prevent overfitting. UNO~\cite{fini2021unified} is a unified objective to handle old and new classes jointly via swapped prediction~\cite{caron2020unsupervised}. NCD assumes all unlabeled data are from new classes.

\noindent\textbf{Generalized Category Discovery} (GCD)~\cite{vaze2022generalized,cao2022openworld} removes the limited assumption and aims to simultaneously cluster old and new classes in the unlabeled data, given some labeled samples of old classes. Pioneer works~\cite{vaze2022generalized,fei2022xcon} conduct supervised~\cite{khosla2020supervised} and unsupervised contrastive learning~\cite{chen2020simple}, and employ semi-supervised K-means~\cite{vaze2022generalized,macqueen1967some} clustering. Later works~\cite{pu2023dynamic,zhang2023promptcal} exploit underlying relationships for better feature representation. Zhao \etal~\cite{Zhao_2023_ICCV} propose an EM-like framework alternating between contrastive learning~\cite{li2021prototypical} and class number estimation. These methods predominantly rely on non-parametric classifiers. By contrast, recent works~\cite{Wen_2023_ICCV,chiaroni2023parametric} propose to avoid prediction biases to achieve remarkable results with parametric classifiers. Although GCD has made great advancements~\cite{NEURIPS2023_3f52ab43,gu2023class}, it inherently suffers from issues like imbalanced accuracy and confidence between old and new classes, which is intractable due to the incompletely transferable knowledge and unlabeled nature of new classes. In this paper, we propose AGCD to address them with affordable labeling budgets.

\noindent\textbf{Active Learning} (AL)~\cite{settles2009active} aims to maximize models' performance with a limited labeling budget. We focus on pool-based AL~\cite{zhan2021comparative}. Sampling strategies contain two types. Uncertainty-based methods select samples with high predictive uncertainty, \eg, entropy~\cite{wang2014new}, least confidence~\cite{wang2014new} and margin~\cite{roth2006margin}. Diversity-based methods select samples that could represent the entire dataset. Typical works include KMeans~\cite{macqueen1967some}, CoreSet~\cite{sener2018active} and BADGE~\cite{Ash2020Deep}. Hybrid methods~\cite{huang2010active,agarwal2020contextual,hsu2015active} combine the two types for further improvements. In principle, AL is in a close-world setting, where labeled and unlabeled data share classes. While in AGCD, unlabeled data contain more categories than labeled data, and models are expected to classify all the classes, not limited to the old classes present in labeled data.

\section{Preliminaries and Analysis}
\label{sec:preliminaries}

Here, we briefly introduce the setting and methods of Generalized Category Discovery (GCD) (Sec.~\ref{subsec:gcd-setting}) and give empirical results to reveal inherent issues (Sec.~\ref{subsec:gcd-problems}), which motivates us to propose our setting AGCD in Sec.~\ref{sec:agcd-setting}.

\vspace{-3pt}
\subsection{Setup and Training Methods of GCD}
\label{subsec:gcd-setting}

\vspace{-2pt}
\paragraph{Problem definition of GCD.}
Given a labeled dataset $\mathcal{D}_l=\{(\mathbf{x}_i^l,y_i^l)\}\subset \mathcal{X}\times\mathcal{Y}_l$ and an unlabeled dataset $\mathcal{D}_u=\{(\mathbf{x}_i^u,y_i^u)\}\subset \mathcal{X}\times\mathcal{Y}_u$. $\mathcal{D}_l$ only contains old classes, while $\mathcal{D}_u$ contains both old and new classes, \ie, $\mathcal{Y}_l=\mathcal{C}_{old}, \mathcal{Y}_u=\mathcal{C}_{old}\cup \mathcal{C}_{new}$. Models are required to cluster both old and new classes in $\mathcal{D}_u$. The number of novel classes $K_{new}$ is known a-prior or estimated~\cite{vaze2022generalized,pu2023dynamic,Zhao_2023_ICCV}. $f(\cdot)$ and $g(\cdot)$ are feature extractor and projection head for contrastive learning respectively. $\mathbf{h}_i=f(\mathbf{x}_i)$ and $\mathbf{z}_i=g(\mathbf{h}_i)$ are $\ell$-2 normalized feature and projected embeddings respectively.

\vspace{-7pt}
\paragraph{Related Training Methods.}
Vaze \etal~\cite{vaze2022generalized} propose to employ supervised~\cite{khosla2020supervised} and self-supervised~\cite{chen2020simple} contrastive learning on labeled $B^l$ and whole mini-batch $B$:
\begin{align}
\vspace{-8pt}
\footnotesize
    \mathcal{L}_{con}^l & = \frac{1}{|B^l|}\sum_{i\in B^l}\frac{1}{|\mathcal{N}(i)|}\sum_{q\in\mathcal{N}(i)}-\log\frac{\exp(\mathbf{z}_i^\top\mathbf{z}_q^\prime/\tau_c)}{\sum_{n\neq i}\exp(\mathbf{z}_i^\top \mathbf{z}_n^\prime/\tau_c)},
    \label{eq:loss-con-l} \\
    \mathcal{L}_{con}^u & = \frac{1}{|B|}\sum_{i\in B}-\log\frac{\exp(\mathbf{z}_i^\top\mathbf{z}_i^\prime/\tau_c)}{\sum_{n\neq i}\exp(\mathbf{z}_i^\top \mathbf{z}_n^\prime/\tau_c)}. \label{eq:loss-con-u}
\end{align}
The overall contrastive loss $\mathcal{L}_{con} = (1-\lambda)\mathcal{L}_{con}^u+\lambda \mathcal{L}_{con}^l$.

SimGCD~\cite{Wen_2023_ICCV} employs a parametric prototypical classifier $\mathcal{C}=\{\mathbf{c}_1,\cdots,\mathbf{c}_K\}$, where $K=K_{old}+K_{new}$. The posterior probability could be expressed as:
\begin{small}
\begin{equation}
\vspace{-5pt}
    \mathbf{p}_i^{(k)}=\frac{\exp(\mathbf{h}_i^\top\mathbf{c}_k)/\tau_p}{\sum_{k^\prime}\exp(\mathbf{h}_i^\top\mathbf{c}_k^\prime)/\tau_p}.
\end{equation}
\end{small}SimGCD implements self-distillation on two views along with an entropy $H(\cdot)$ regularization across all samples:
\begin{equation}
    \mathcal{L}_{cls}^u = \frac{1}{|B|}\ell(\mathbf{q}_i^\prime,\mathbf{p}_i)-\lambda_e H(\overline{\mathbf{p}}),
    \label{eq:loss-simgcd-cls-u}
    \vspace{-5pt}
\end{equation}
where $\mathbf{q}_i^\prime$ is a sharpened probability of another view, and $\overline{\mathbf{p}}=\frac{1}{2|B|}\sum_{i\in B}(\mathbf{p}_i+\mathbf{p}_i^\prime)$, $\ell(\cdot)$ denotes cross-entropy loss. The supervised loss is also employed on $\mathcal{D}_l$ with labels $y_i$:
\begin{equation}
\vspace{-5pt}
    \mathcal{L}_{cls}^l = \frac{1}{|B^l|}\sum_{i\in B^l}\ell(y_i,\mathbf{p}_i).
    \label{eq:loss-simgcd-cls-l}
\end{equation}

\subsection{Inherent Problems in GCD}
\label{subsec:gcd-problems}

SimGCD~\cite{Wen_2023_ICCV} is the state-of-the-art (SOTA) and effective GCD method with a parametric classifier, we thus use it for the analysis of confidence~\cite{guo2017calibration} and accuracy. To find the problems of GCD in real scenarios, we train models with SimGCD in a practical low-label condition as in Table~\ref{tab:datasets-settings}.

\begin{figure}[!htb]
    \centering
    \includegraphics[width=.75\linewidth]{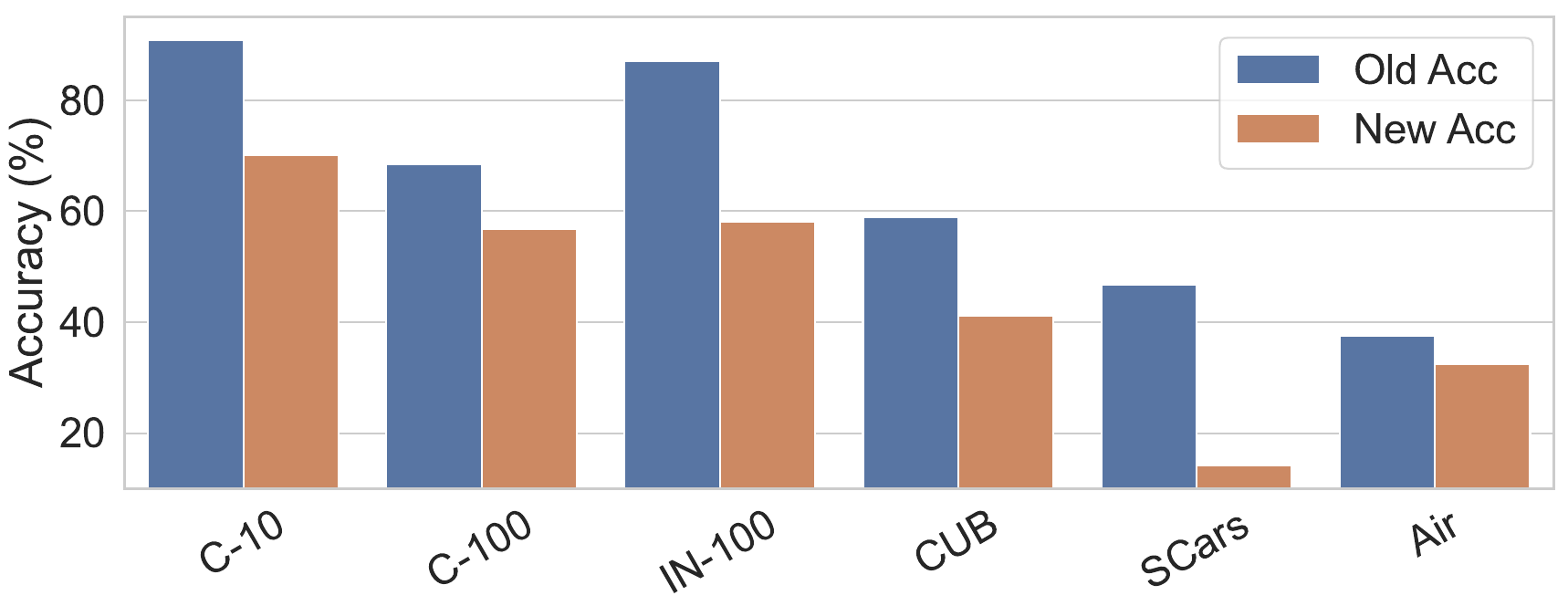}
    \vspace{-9pt}
    \caption{Accuracy of old and new classes on six datasets.}
    \label{fig:old-new-acc-imbalance}
    \vspace{-13pt}
\end{figure}

\vspace{-7pt}
\paragraph{GCD suffers from severe performance mismatch between old and new classes.}
As Fig.~\ref{fig:old-new-acc-imbalance} shows, the accuracy of old classes largely surpasses the new ones, which is imbalanced. For example, the performance gap is 28.88\% and 32.59\% on IN-100 and SCars. The underlying reason is the inherent label condition imbalanced, \ie, old classes are partially labeled while new classes are fully unlabeled, which is the essential issue of the GCD task itself.

\begin{figure}[!htb]
    \centering
    \includegraphics[width=.9\linewidth]{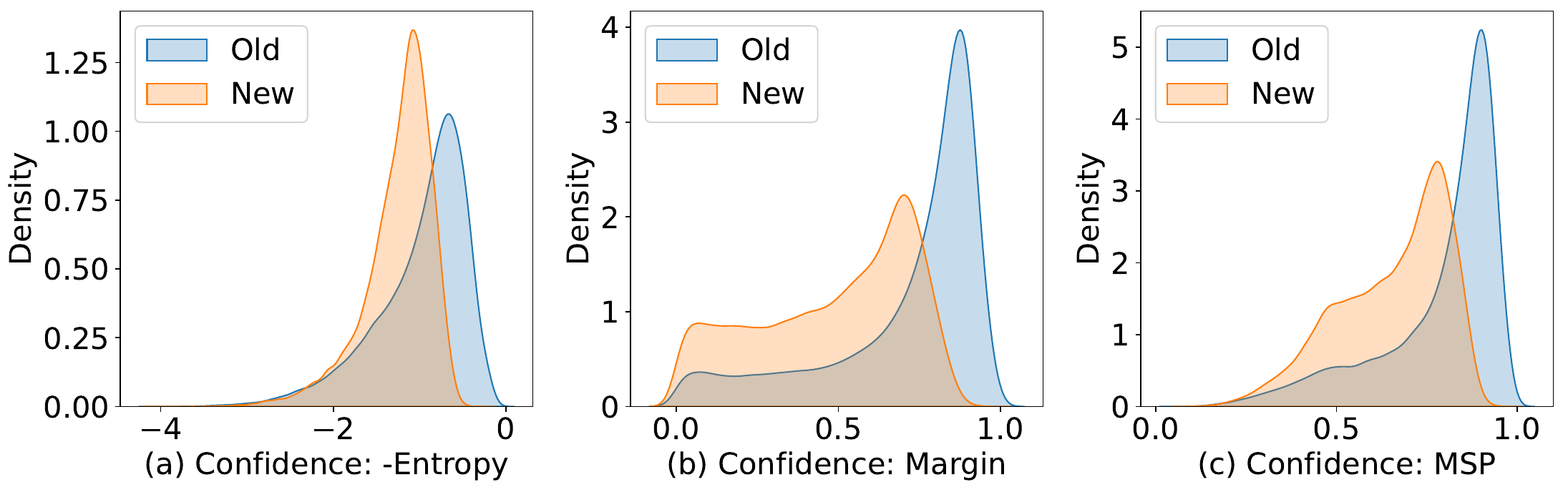}
    \vspace{-7pt}
    \caption{Confidence distribution on ImageNet-100. Three measures: -entropy (a), margin (b) and MSP (c).}
    \label{fig:old-new-confidence}
    \vspace{-13pt}
\end{figure}

\begin{figure*}[!t]
    \centering
    \includegraphics[width=0.98\linewidth]{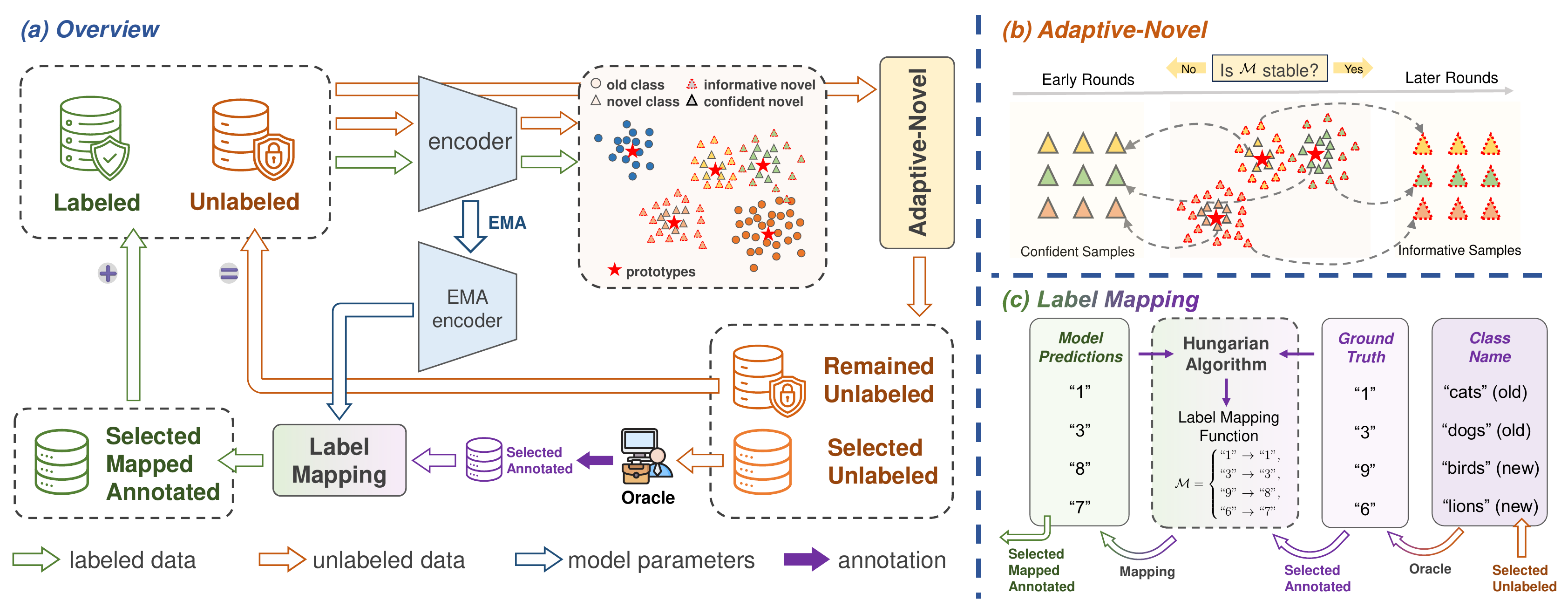}
    \vspace{-10pt}
    \caption{The framework of AGCD. (a) Overall pipeline and dataflow. Models are trained on $\mathcal{D}_l^t\cup\mathcal{D}_u^t$ with SimGCD, and select samples in $\mathcal{D}_u^t$. (b) The proposed \texttt{Adaptive-Novel} sampling strategy. Here $\mathcal{M}$ denotes the \emph{label mapping function}. Stable $\mathcal{M}$ means that at the initial and end epochs of the current round, $\mathcal{M}$ does not change largely. \emph{Confident novel samples} are sampled at early rounds, when $\mathcal{M}$ is stable, we select the \emph{informative} ones. (c) Illustration of \emph{label mapping} computed by model predictions and ground truth on $\mathcal{D}_l^{t-1}\cup\mathcal{D}_q^t$.}
    \vspace{-10pt}
    \label{fig:agcd-pipeline}
\end{figure*}

\vspace{-7pt}
\paragraph{Models tend to have inconsistent confidence between old and new classes.}
We plot the distribution of the model's predictive confidence, with three metrics, \ie, (minus) entropy~\cite{shannon2001mathematical}, margin~\cite{roth2006margin} and maximum softmax probability (MSP)~\cite{hendrycks2017a}. From Fig.~\ref{fig:old-new-confidence}, we observe that the confidence distribution between old and new classes is inconsistent. The confidence of old classes is relatively high, which aligns with intuition because some old samples are labeled, while novel classes are learned with soft targets in Eq.~\eqref{eq:loss-simgcd-cls-u}, resulting in ambiguous predictions.

\section{Active Generalized Category Discovery}
\label{sec:agcd-setting}

\vspace{-5pt}
As in Sec.~\ref{subsec:gcd-problems}, models perform poorly on the fully unlabeled new classes across various datasets and are unable to correct errors. We argue that some annotations on new classes are deemed indispensable. To avoid excessive labeling costs and make the setting practical, we aim to enhance GCD with limited annotation budgets and propose Active Generalized Category Discovery (AGCD). Firstly, we introduce the problem definition, analyze its challenges and distinguish the differences from AL (Sec.~\ref{subsec:agcd-task-analysis}). Then, we provide metrics considering the accuracy and novelty of samples (Sec.~\ref{subsec:evaluation-metrics}). Next, we elaborate on the proposed sampling strategy \texttt{Adaptive-Novel} (Sec.~\ref{subsec:adaptive-novel-sampling}) and stable label mapping algorithm (Sec.~\ref{subsec:stable-label-mapping}).

\vspace{-13pt}
\paragraph{Overview.}
We provide the framework of AGCD in Fig.~\ref{fig:agcd-pipeline}. (a) illustrates the pipeline and dataflow of AGCD, and models are trained with off-the-shelf SimGCD~\cite{Wen_2023_ICCV}. (b) shows the proposed \texttt{Adaptive-Novel} query strategy. (c) demonstrates the \emph{labeling mapping} method to transform ground truth labels to models' label space.

\subsection{The Task of AGCD and Basic Analysis}
\label{subsec:agcd-task-analysis}

\vspace{-3pt}
\paragraph{Problem definition of AGCD.}
Initially, the model is trained on both $\mathcal{D}_l^{init}=\mathcal{D}_l^0=\{(\mathbf{x}_i^l,y_i^l)\}\subset \mathcal{X}\times\mathcal{Y}_l$ and $\mathcal{D}_u^{init}=\mathcal{D}_u^0=\{(\mathbf{x}_i^u,y_i^u)\}\subset \mathcal{X}\times\mathcal{Y}_u$ with off-the-shelf GCD training method SimGCD~\cite{Wen_2023_ICCV}. The initial data splits are similar to GCD~\cite{vaze2022generalized} as in Table~\ref{tab:datasets-settings}. After this base training stage, AGCD could have multiple rounds (denoted as $n$ in total) like AL. At round $t$, the model first selects a batch of $b$ samples (budget size) from unlabeled data $\mathcal{D}_u^{t-1}$ and queries its labels to obtain $\mathcal{D}_q^t=\{(\mathbf{x}_i^q,y_i^q)\}$. Then labeled and unlabeled data are updated as $\mathcal{D}_l^t=\mathcal{D}_l^{t-1}\cup\mathcal{D}_q^t\subset\mathcal{X}\times\mathcal{Y}_l^t$, $\mathcal{D}_u^t=\mathcal{D}_u^{t-1}\setminus\mathcal{D}_q^t$. Models are then trained on $\mathcal{D}_l^t\cup \mathcal{D}_u^t$ with off-the-shelf SimGCD~\cite{Wen_2023_ICCV}. Note that initial labeled data only contains old classes, \ie $\mathcal{Y}_l^0=\mathcal{C}_{old}$, but after querying in AGCD, the labeled data $\mathcal{D}_l^t$ could contain some classes of all novel ones $\mathcal{C}_{new}$. The total budget size is $b\times n$. For the queried data $\mathcal{D}_q^t$, models are trained with supervised loss $\mathcal{L}_{con}^l$ in Eq.~\eqref{eq:loss-con-l} and $\mathcal{L}_{cls}^l$ in Eq.~\eqref{eq:loss-simgcd-cls-l}.

\begin{table}[!t]
\setlength\tabcolsep{4pt}
\centering
\renewcommand{\arraystretch}{1}
\caption{The default setting of 3 \colorbox{color2}{generic datasets} and 3 \colorbox{color3}{fine-grained datasets} in AGCD benchmark. $|\mathcal{Y}_l^\text{init}|=K_{old}$, $|\mathcal{Y}_u^\text{init}|=K_{old}+K_{new}$ denote the initial number of classes in $|\mathcal{D}_l^\text{init}|$ and $|\mathcal{D}_u^\text{init}|$. The number of queries across all rounds is displayed in both the total count and average count per class.}
\vspace{-8pt}
\label{tab:datasets-settings}
\resizebox{.98\linewidth}{!}{
\begin{tabular}{@{}cccccccc@{}}
    \toprule
    \multirow{2}{*}{Dataset} & \multicolumn{2}{c}{Labeled $\mathcal{D}_l^\text{init}$} & \multicolumn{2}{c}{Unlabeled $\mathcal{D}_u^\text{init}$} & \multirow{2}{*}{\#Rounds} & \multirow{2}{*}{\makecell[c]{\#Query\\(total)}} & \multirow{2}{*}{\makecell[c]{\#Query\\(per class)}} \\ \cmidrule(lr){2-3} \cmidrule(lr){4-5}
     & $|\mathcal{D}_l^\text{init}|$ & $|\mathcal{Y}_l^\text{init}|$ & $|\mathcal{D}_u^\text{init}|$ & $|\mathcal{Y}_u^\text{init}|$ &  &  &  \\ \midrule
    \cellcolor{color2}CIFAR10 (C-10)~\cite{krizhevsky2009learning} & 2,000 & 2 & 48,000 & 10 & 1 & 100 & 10 \\
    \cellcolor{color2}CIFAR100 (C-100)~\cite{krizhevsky2009learning} & 5,000 & 50 & 45,000 & 100 & 5 & 500 & 5 \\
    \cellcolor{color2}ImageNet-100 (IN-100)~\cite{deng2009imagenet} & 12,744 & 50 & 114,371 & 100 & 5 & 500 & 5 \\ \midrule
    \cellcolor{color3}CUB (CUB)~\cite{wah2011caltech} & 599 & 100 & 5,395 & 200 & 5 & 500 & 2.5 \\
    \cellcolor{color3}Stanford Cars (SCars)~\cite{krause20133d} & 800 & 98 & 7,344 & 196 & 5 & 500 & 2.5 \\
    \cellcolor{color3}FGVC-Aircraft (Air)~\cite{maji2013fine} & 666 & 50 & 6,001 & 100 & 5 & 500 & 5 \\ \bottomrule
\end{tabular}
}
\vspace{-10pt}
\end{table}

\vspace{-13pt}
\paragraph{Two challenges in AGCD.}
(1) Directly employing conventional AL methods (\eg, Entropy) results in sub-optimal performance. This is because they do not consider new classes, and the confidence and feature distribution of old and new classes are inconsistent, as discussed in Sec.~\ref{subsec:gcd-problems}. (2) Considering the clustering of GCD, the queried labels could not be directly sent to models due to the different ordering of indices of new classes. For example, considering new classes ``birds'' and ``lions'', the model assigns ``8'', ``7'' to them while the ground truth is ``9'', ``6'', then we should obtain a mapping as: ``9''$\to$``8'', ``6''$\to$``7'', and map the ground truth to the model's label space for supervision.

\vspace{-13pt}
\paragraph{Distinguishing between AL and AGCD.} (1) AGCD could be viewed as an open-world extrapolated version of AL requiring models to classify both old and new classes, and the unlabeled data could contain new classes. (2) In conventional AL, models are not trained on $\mathcal{D}_u$, which is only used for sample selection and only the selected samples engage in training. In contrast, in AGCD, models not only select samples in $\mathcal{D}_u$ but are also trained on it.

\subsection{Evaluation and Metrics}
\label{subsec:evaluation-metrics}

\paragraph{Accuracy Evaluation.}
GCD adopts a \emph{transductive} evaluation on unlabeled training data $\mathcal{D}_u$. By contrast, we adopt an \emph{inductive} evaluation for AGCD, \ie, we test models on the unseen and disjoint test dataset. The reason is that models query some labels in $\mathcal{D}_u$, making it unfair for evaluation. The accuracy is calculated using ground truth labels $y_i$ and models' predictions $\hat y_i$ as follows:
\begin{equation}
\vspace{-5pt}
    ACC=\max_{p\in \mathcal{P}(\mathcal{Y}_u)} \frac{1}{M}\sum_{i=1}^M \mathds{1}(y_i=p(\hat{y}_i)),
    \label{eq:gcd-acc-evaluation}
\end{equation}
$M=|\mathcal{D}_u|$ and $\mathcal{P}(\mathcal{Y}_u)$ is the set of all permutations across all classes $\mathcal{C}_{old}\cup\mathcal{C}_{new}$. The maximum value is computed by the Hungarian optimal assignment algorithm~\cite{kuhn1955hungarian} across all $(K_{old}+K_{new})$ classes, which is the same as GCD~\cite{vaze2022generalized}.

\vspace{-10pt}
\paragraph{Novelty Evaluation.}
We also evaluate the selected samples' category attribution with the following metrics: (1) Novelty Coverage $\texttt{Nov-C}$ that measures the coverage of new classes. (2) Novelty Ratio $\texttt{Nov-R}$ that measures the ratio of the selected samples belonging to new classes. (3) Novelty Uniformity $\texttt{Nov-U}$ that measures the uniformity of the coverage across new classes. (4) Novelty Information $\texttt{Nov-I}$, which considers both ratio and uniformity, a high value indicates one could neither randomly select samples across old and new classes nor from very few new classes. Specifically, these metrics are formulated as:
\vspace{-5pt}
\begin{small}
\begin{align}
    \texttt{Nov-C} & = |\mathcal{C}_{new,select}|/K_{new}, \\
    \texttt{Nov-R} & = \sum_{i=1}^{N_{select}}\frac{\mathds{1}(y_i\in\mathcal{C}_{new})}{N_{select}}, \\
    \texttt{Nov-U} & = -\sum_{c=1}^{K_{new}}\frac{N_{new,i}}{N_{select}}\log \frac{N_{new,i}}{N_{select}}/\log K_{new}, \\
    \texttt{Nov-I} & = \texttt{Nov-R}\times\texttt{Nov-U},
\end{align}
\end{small}where $N_{select}$ and $N_{new,i}$ are the numbers of samples in total and those belonging to the $i$-th new class respectively. $\mathcal{C}_{new,select}$ denotes the selected new classes.

\subsection{Adaptive Novel Sampling}
\label{subsec:adaptive-novel-sampling}

In AGCD, we simultaneously consider three aspects: \emph{novelty}, \emph{informativeness} and \emph{diversity} of samples and propose an adaptive sampling strategy called \texttt{Adaptive-Novel}, as in Fig.~\ref{fig:agcd-pipeline} (b). (1) For the aspect of novelty, as the initial labeling condition is severely imbalanced, we should give priority to selecting samples from new classes. Models' predictions $\hat y_i$ are proxies of samples' novelty. (2) For the aspect of diversity, we uniformly select samples from novel classes, \ie, at each round, we select $\lfloor b/K_{new}\rfloor$ samples in each new class based on the model's prediction. (3) For the aspect of informativeness, we choose Margin~\cite{roth2006margin} as the uncertainty metric. Selecting the most uncertain or informative samples~\cite{scheffer2001active,settles2009active} has been a consensus in the literature of AL. However, new classes are initially fully unlabeled, and clusters of new classes might be unstable and biased~\cite{arazo2020pseudo}, please refer to visualizations of the appendix, thus including difficult samples at early rounds to biased clusters hinders training. Here, we name two types of samples of novel classes, the most ambiguous and informative samples are named \emph{informative novel samples} while the most certain ones are called \emph{confident novel samples}. We devise an adaptive mechanism, where models select \emph{confident novel samples} with minimum uncertainties to rectify and stabilize novel clusters at initial rounds. While at later rounds, \emph{informative novel samples} with maximum uncertainties are selected to refine decision boundaries and further improve the performance.

In our method, we are expected to capture which type of samples are more important to the current model. To achieve this goal, we offer a heuristic criterion, where we regard the stability of label mapping $\mathcal{M}$ (as in Sec.~\ref{subsec:stable-label-mapping}) between model predictions and ground truth as the stability of the clusters, especially for new classes. If at round $k$,  the change of $\mathcal{M}$ between the start and end epochs is negligible, the clustering is deemed stable and we can transfer to sample \emph{informative novel samples} from round $k+1$.

\begin{table*}[!t]
\setlength\tabcolsep{6pt}
\centering
\renewcommand{\arraystretch}{1}
\caption{Comparative results of various methods with 5 rounds of active category discovery on generic datasets. Our method outperforms several uncertainty-based (Unc.) and representative/diversity-based (Rep./Div.) methods. Mean results over three runs are reported.}
\vspace{-7pt}
\label{tab:main-acc-generic}
\resizebox{.85\linewidth}{!}{
\begin{tabular}{@{}ccccccccccc@{}}
    \toprule
    \multirow{2}{*}{Type} & \multirow{2}{*}{AL Strategies} & \multicolumn{3}{c}{CIFAR10} & \multicolumn{3}{c}{CIFAR100} & \multicolumn{3}{c}{ImageNet-100} \\ \cmidrule(l){3-5} \cmidrule(l){6-8} \cmidrule(l){9-11} 
     &  & All & Old & New & All & Old & New & All & Old & New \\ \midrule
    \multirow{2}{*}{Baseline} & w/o AGCD & 74.22 & 90.80 & 70.07 & 62.62 & 68.46 & 56.78 & 72.56 & 87.00 & 58.12 \\
     & \texttt{Random} & 82.74 & 93.05 & 80.16 & 67.28 & 74.52 & 60.04 & 79.16 & 89.40 & 68.92 \\ \midrule \midrule
    \multirow{3}{*}{Unc.} & \texttt{Entropy}~\cite{wang2014new} & 76.25 & 95.55 & 71.43 & 64.59 & 73.94 & 55.24 & 75.96 & 91.04 & 60.88 \\
     & \texttt{LeastConf}~\cite{wang2014new} & 78.32 & \textbf{96.00} & 73.90 & 65.63 & \textbf{76.74} & 54.52 & 76.82 & 91.92 & 61.72 \\
     & \texttt{Margin}~\cite{roth2006margin} & 92.34 & 94.35 & 91.84 & 69.08 & 75.58 & 62.58 & 80.46 & 92.40 & 68.52 \\ \midrule \midrule
    \multirow{3}{*}{Rep./Div.} & \texttt{KMeans}~\cite{macqueen1967some} & 91.18 & 93.10 & 90.70 & 66.70 & 72.66 & 60.74 & 78.18 & 90.08 & 66.28 \\
     & \texttt{CoreSet}~\cite{sener2018active} & 85.51 & 94.95 & 83.15 & 65.72 & 77.64 & 53.80 & 78.08 & 91.92 & 64.24 \\
     & \texttt{BADGE}~\cite{Ash2020Deep} & 92.31 & 94.75 & 91.70 & 67.22 & 73.70 & 60.74 & 81.48 & \textbf{92.68} & 70.28 \\ \midrule \midrule
    \RC{30}Ours & \texttt{Adaptive-Novel} & \textbf{93.15} & 94.55 & \textbf{92.80} & \textbf{71.25} & 75.72 & \textbf{66.78} & \textbf{83.34} & 90.20 & \textbf{76.48} \\ \bottomrule
\end{tabular}
}
\end{table*}

\subsection{Stable Label Mapping Algorithm}
\label{subsec:stable-label-mapping}

As the queried labels could not be directly used, one should perform \emph{label mapping} to ``translate'' ground truth labels to the model's label space. as in Fig.~\ref{fig:agcd-pipeline} (c). We propose to calculate the mapping function from ground truth to the model's perspective via Hungarian algorithm~\cite{kuhn1955hungarian}, similar to Eq.~\eqref{eq:gcd-acc-evaluation}. However, we can only perform label mapping using accessible labeled data, \ie, $\mathcal{D}_l^{t}=\mathcal{D}_l^{t-1}\cup\mathcal{D}_q^t$ at round $t$, which is very limited, especially for new classes, and could bring about unstable results. To alleviate this, we maintain an exponential moving average (EMA)~\cite{tarvainen2017mean,zhu2022rethinking,NEURIPS2023_98143953} of the model, and compute the \emph{label mapping function} utilizing the EMA model's predictions on $\mathcal{D}_l^t=\mathcal{D}_l^{t-1}\cup\mathcal{D}_q^t$:
\begin{equation}
\vspace{-7pt}
    \mathcal{M}^t=\argmax_{m\in \mathcal{P}(\mathcal{C}_{all})} \frac{1}{|\mathcal{D}_l^t|}\sum_{i\in \mathcal{D}_l^t} \mathds{1}(m(y_i)=\hat{y}^{ema}_i),
    \label{eq:label-mapping}
\end{equation}
where $y_i$ and $\hat y_i^{ema}$ are ground truth and predicted labels by EMA models. $\mathcal{P}(\mathcal{C}_{all})$ is the permutation across all classes $\mathcal{C}_{old}\cup\mathcal{C}_{new}$. $\mathcal{M}^t$ is a one-to-one mapping between two sets of classes. Then the mapped label of each query is $y_i^{map}=\mathcal{M}(y_i)$. Let $\mathcal{D}_{q,map}^t$ denote the query dataset after \emph{label mapping}, the mapped labeled data is $\mathcal{D}_l^t=\mathcal{D}_l^{t-1}\cup\mathcal{D}_{q,map}^t$.

\section{Experiments}
\label{sec:experiments}

\subsection{Experimental Setup}

\begin{table*}[!t]
\setlength\tabcolsep{6pt}
\centering
\renewcommand{\arraystretch}{1}
\caption{Comparative results of various methods with 5 rounds of active category discovery on fine-grained datasets. Our method outperforms several uncertainty-based (Unc.) and representative/diversity-based (Rep./Div.) methods. Mean results over three runs are reported.}
\vspace{-7pt}
\label{tab:main-acc-fine}
\resizebox{.85\linewidth}{!}{
\begin{tabular}{@{}ccccccccccc@{}}
    \toprule
    \multirow{2}{*}{Type} & \multirow{2}{*}{Query Strategies} & \multicolumn{3}{c}{CUB} & \multicolumn{3}{c}{Stanford Cars} & \multicolumn{3}{c}{FGVC-Aircraft} \\ \cmidrule(l){3-5} \cmidrule(l){6-8} \cmidrule(l){9-11} 
     &  & All & Old & New & All & Old & New & All & Old & New \\ \midrule
    \multirow{2}{*}{Baseline} & w/o AGCD & 50.17 & 58.95 & 41.18 & 30.12 & 46.71 & 14.12 & 35.01 & 37.53 & 32.49 \\
     & \texttt{Random} & 62.74 & 64.88 & 60.62 & 44.12 & 53.44 & 35.13 & 50.41 & 51.38 & 49.43 \\ \midrule \midrule
    \multirow{3}{*}{Unc.} & \texttt{Entropy}~\cite{wang2014new} & 62.82 & \textbf{69.52} & 56.19 & 42.40 & 53.75 & 31.44 & 43.89 & 51.92 & 35.86 \\
     & \texttt{LeastConf}~\cite{wang2014new} & 61.48 & 66.12 & 56.87 & 45.82 & 55.32 & 36.65 & 44.91 & 50.42 & 39.40 \\
     & \texttt{Margin}~\cite{roth2006margin} & 65.08 & 68.41 & 61.79 & 46.03 & 57.67 & 34.79 & 51.37 & \textbf{52.46} & 50.27 \\ \midrule \midrule
    \multirow{3}{*}{Rep./Div.} & \texttt{KMeans}~\cite{macqueen1967some} & 61.30 & 68.27 & 54.40 & 40.79 & 52.99 & 29.03 & 51.58 & 51.08 & 52.07 \\
     & \texttt{CoreSet}~\cite{sener2018active} & 63.44 & 65.95 & 60.96 & 42.52 & 52.00 & 33.37 & 45.03 & 51.68 & 38.38 \\
     & \texttt{BADGE}~\cite{Ash2020Deep} & 65.84 & 69.00 & 62.71 & 45.82 & 54.41 & 37.53 & 52.03 & 51.68 & 52.37 \\ \midrule \midrule
    \RC{30}Ours & \texttt{Adaptive-Novel} & \textbf{66.62} & 66.54 & \textbf{66.70} & \textbf{48.36} & \textbf{57.73} & \textbf{39.34} & \textbf{53.74} & 51.50 & \textbf{55.98} \\ \bottomrule
\end{tabular}
}
\end{table*}

\paragraph{Datasets.}
We construct AGCD on six datasets as shown in Table~\ref{tab:datasets-settings}. For each dataset, $K_{old}$ classes are selected as ``old'' classes, while the remaining $K_{new}$ classes are ``new'' classes. We then sub-sample 20\% of the training samples in $K_{old}$ as the initial labeled dataset $\mathcal{D}_l^\text{init}$, while all remaining samples constitute the initial unlabeled part $\mathcal{D}_u^\text{init}$ for querying in subsequent rounds. The construction of $\mathcal{D}_l^\text{init}$ and $\mathcal{D}_u^\text{init}$ is similar to the literature of GCD~\cite{vaze2022generalized,Wen_2023_ICCV}, but with fewer labeling ratio which is closer to a real-world scenario.

\vspace{-10pt}
\paragraph{Evaluation.}
We compare the accuracy of AGCD with various query strategies. For fair comparisons, we use off-the-shelf SimGCD~\cite{Wen_2023_ICCV} for training as SimGCD is effective and the SOTA method in GCD. We employ model EMA for all query methods. Models are evaluated on disjoint test data using Eq.~\eqref{eq:gcd-acc-evaluation} in an \emph{inductive} setting. 

\vspace{-13pt}
\paragraph{Query strategies for comparison.}
We compare our method \texttt{Adaptive-Novel} with various AL strategies~\cite{settles2009active,ren2021survey,zhan2022comparative}, \eg, Random Sampling (\texttt{Random}), uncertainty-based and representative/diversity-based sampling methods. For the uncertainty-based methods, we compare Maximum Entropy (\texttt{Entropy})~\cite{wang2014new}, and Least Confidence (\texttt{LeastConf})~\cite{wang2014new}, Least Margin (\texttt{Margin})~\cite{roth2006margin}. The representative-based methods include KMeans Clustering (\texttt{KMeans})~\cite{macqueen1967some}, Core-Set (\texttt{CoreSet})~\cite{sener2018active}, and Batch Active learning by Diverse Gradient Embeddings (\texttt{BADGE})~\cite{Ash2020Deep}. More details are in the Appendix.

\vspace{-10pt}
\paragraph{AGCD pipeline and implementation details.}
Following GCD~\cite{vaze2022generalized,Wen_2023_ICCV}, we employ ViT-B/16~\cite{dosovitskiy2021an} pre-trained by DINO~\cite{caron2021emerging} as the backbone, and fine-tune only the last transformer block for all experiments. The output of $\texttt{[CLS]}$ token is chosen as the feature representation. The batch size for the original dataset $\mathcal{D}_l$ and $\mathcal{D}_u$ is 128. For queried samples, we use a smaller batch size $\mathcal{B}_q=8$. We implement the base training stage like GCD for 200 epochs and choose models as initialization for AGCD. At each round, we train models on $\mathcal{D}_l^t$ and $\mathcal{D}_u^t$ by various query strategies for 15 epochs. All selection methods are trained using SimGCD with an initial learning rate of $0.1$ and a cosine annealed schedule both in the base training stage and subsequent AGCD stage. All experiments are conducted on NVIDIA RTX A6000 GPUs. More details are in the appendix.

\subsection{Comparative Results}

\paragraph{\texttt{Adaptive-Novel} achieves stronger overall performance.}
As in Table~\ref{tab:main-acc-generic} and Table~\ref{tab:main-acc-fine}, our method outperforms others consistently on various generic and fine-grained datasets. For example, on CIFAR100, our method outperforms \texttt{Random} by 3.97\%/6.74\%, and \texttt{CoreSet} by 5.53\%/12.98\% in terms of accuracy of all/new classes. Overall, our method exhibits an obvious advantage, especially in the accuracy of new classes.

\vspace{-10pt}
\paragraph{\texttt{Adaptive-Novel} achieves more balanced results between old and new classes.}
For all six datasets, the difference in accuracy between old and new classes of our method is minimal, indicating that our method effectively addresses the imbalanced issue in Sec.~\ref{subsec:gcd-problems}. One of the key insights is to prioritize samples from new classes for annotation, which helps to alleviate the inherent imbalanced labeling condition of GCD. For example in CUB, the divergence of old and new accuracy is 0.16\%, while for other methods ranging from $\sim$ 4\% to $\sim$ 14\%. And in ImageNet-100, our method reduces the gap from $\sim$ 28\% to 13.72\%.

\vspace{-10pt}
\paragraph{\texttt{Adaptive-Novel} significantly improves GCD with a limited budget size.}
As in Table~\ref{tab:main-acc-fine}, when selecting only $\sim 2.5$ samples per class for annotation, the accuracy of new classes improves by 25.52\%/25.22\% on CUB/SCars, showcasing the efficiency and practicality of AGCD.

\vspace{-5pt}
\begin{table}[!t]
\setlength\tabcolsep{2pt}
\centering
\renewcommand{\arraystretch}{1}
\caption{Novelty metrics of all the selected data over 5 rounds on CIFAR100 and Stanford Cars.}
\vspace{-7pt}
\label{tab:novelty-c100-cars}
\resizebox{1.0\linewidth}{!}{
\begin{tabular}{@{}ccccccccc@{}}
    \toprule
    \multirow{2}{*}{AL Strategies} & \multicolumn{4}{c}{CIFAR100} & \multicolumn{4}{c}{Stanford Cars} \\ \cmidrule(l){2-5} \cmidrule(l){6-9} 
     & \texttt{Nov-C} & \texttt{Nov-R} & \texttt{Nov-U} & \texttt{Nov-I} & \texttt{Nov-C} & \texttt{Nov-R} & \texttt{Nov-U} & \texttt{Nov-I} \\ \midrule
    \texttt{Random} & \textbf{1.00} & 0.52 & 0.97 & 0.50 & 0.93 & 0.57 & 0.96 & 0.55 \\
    \texttt{Entropy} & 0.90 & 0.44 & 0.91 & 0.40 & 0.85 & 0.64 & 0.92 & 0.59 \\
    \texttt{Margin} & 0.96 & 0.63 & 0.95 & 0.60 & 0.90 & 0.66 & 0.93 & 0.61 \\
    \texttt{CoreSet} & 0.96 & 0.61 & 0.94 & 0.57 & 0.89 & \textbf{0.69} & 0.94 & 0.65 \\
    \texttt{BADGE} & \textbf{1.00} & 0.63 & \textbf{0.98} & 0.62 & 0.95 & 0.64 & \textbf{0.97} & 0.62 \\ \midrule
    \RC{30}Ours & \textbf{1.00} & \textbf{0.71} & \textbf{0.98} & \textbf{0.70} & \textbf{0.98} & \textbf{0.69} & \textbf{0.97} & \textbf{0.67} \\ \bottomrule
\end{tabular}
}
\vspace{-5pt}
\end{table}

\vspace{-5pt}
\paragraph{Do more novel samples necessarily lead to better performance?}
By comparing the results in Table~\ref{tab:main-acc-generic}, Table~\ref{tab:main-acc-fine} and Table~\ref{tab:novelty-c100-cars}, our method generally samples more novel samples with more comprehensive class coverage, which contributes to the remarkable results. However, on Scars, \texttt{BADGE} selects more novel samples than \texttt{Margin}, but its performance is worse. As a result, solely sampling more novel class samples does not necessarily work fine, it is also important to consider the value of different samples.

\begin{table}[!t]
\setlength\tabcolsep{3pt}
\centering
\renewcommand{\arraystretch}{1}
\caption{Ablations on three key factors, \ie, novelty, informativeness and diversity for sample selection in AGCD.}
\vspace{-7pt}
\label{tab:ablation-three-factors}
\resizebox{1.0\linewidth}{!}{
\begin{tabular}{@{}cccccccccc@{}}
    \toprule
    \multirow{2}{*}{ID} & \multirow{2}{*}{Novelty} & \multirow{2}{*}{Informativeness} & \multirow{2}{*}{Diversity} & \multicolumn{3}{c}{CIFAR100} & \multicolumn{3}{c}{CUB} \\ \cmidrule(l){5-7} \cmidrule(l){8-10} 
     &  &  &  & All & Old & New & All & Old & New \\ \midrule
    (a) & \xmark & \xmark & \xmark & 67.28 & 74.52 & 60.04 & 62.74 & 64.88 & 60.62 \\
    (b) & \cmark & \xmark & \xmark & 69.33 & 72.76 & 65.90 & 63.58 & 63.31 & 63.85 \\
    (c) & \cmark & \cmark & \xmark & 69.65 & 75.56 & 63.74 & 64.28 & 65.33 & 63.23 \\
    (d) & \cmark & \cmark & \cmark & \textbf{71.25} & \textbf{75.72} & \textbf{66.78} & \textbf{66.62} & \textbf{66.54} & \textbf{66.70} \\ \bottomrule
\end{tabular}
}
\vspace{-5pt}
\end{table}

\subsection{Ablation Studies}
\label{subsec:ablations}

\begin{figure}[!ht]
    \centering
    \includegraphics[width=0.95\linewidth]{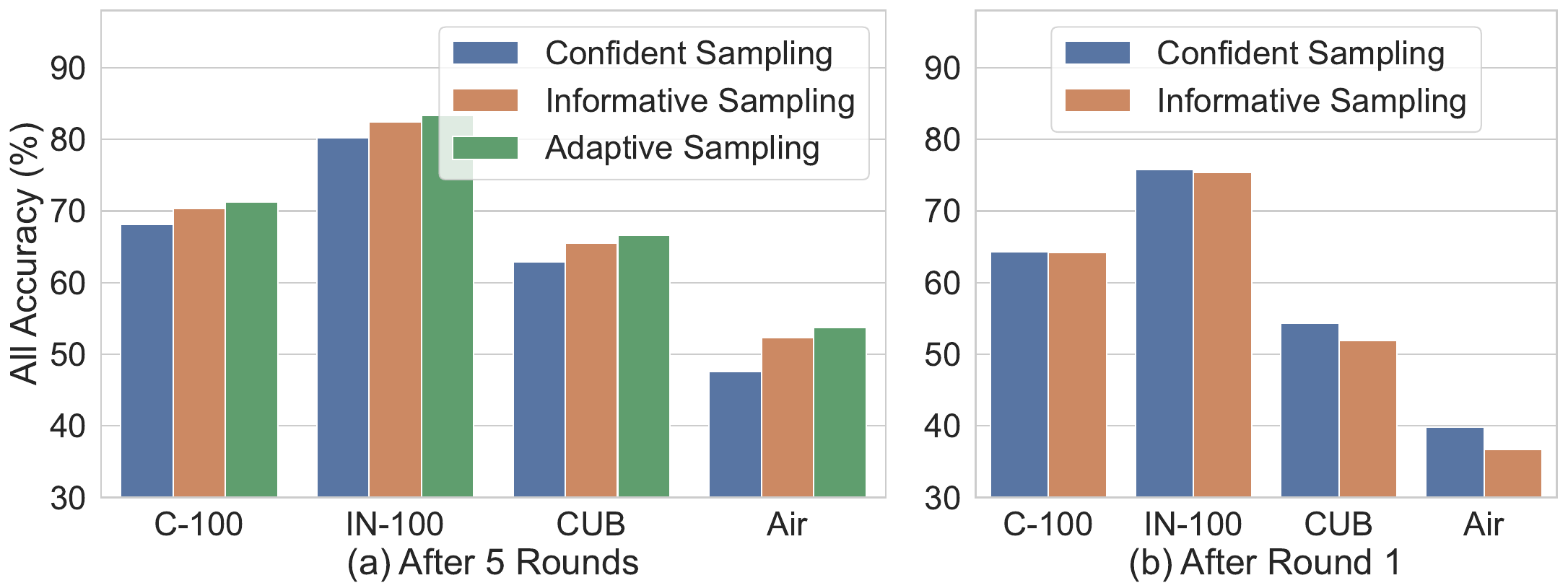}
    \vspace{-7pt}
    \caption{Ablation on adaptive sampling.}
    \label{fig:ablation-adaptive}
    \vspace{-7pt}
\end{figure}

\begin{figure}[!t]
    \centering
    \includegraphics[width=0.95\linewidth]{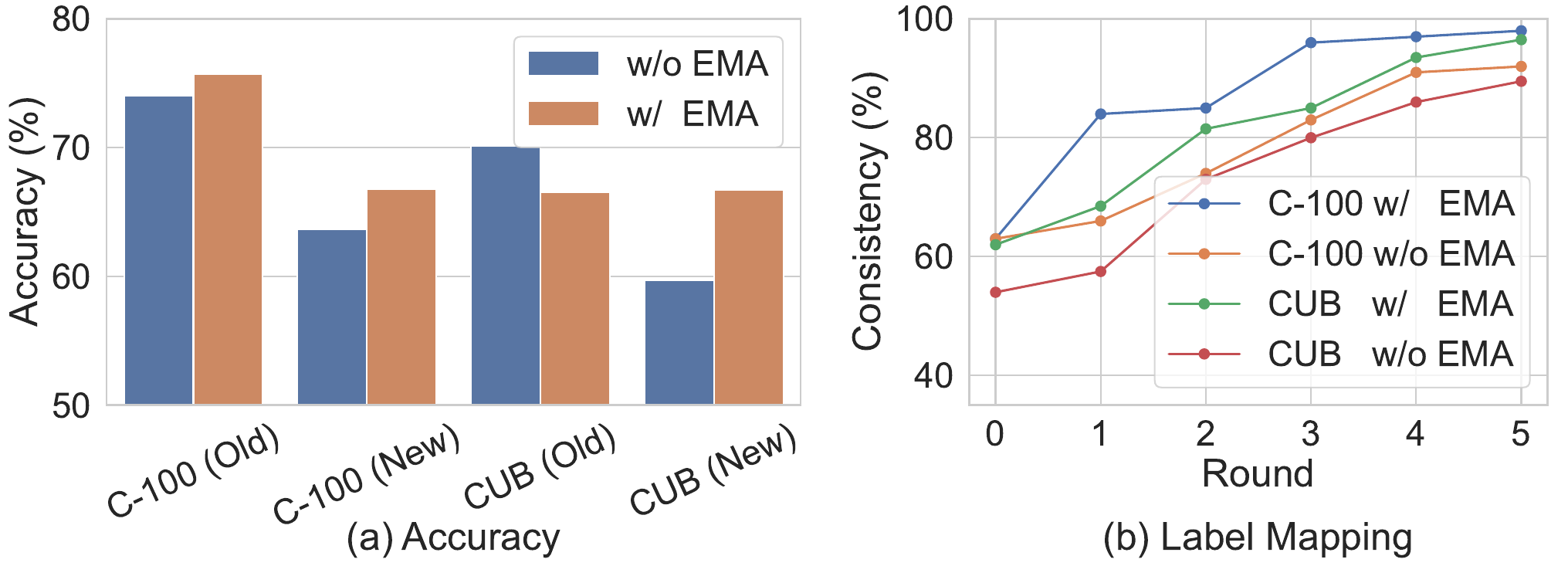}
    \vspace{-7pt}
    \caption{Ablation on model EMA.}
    \label{fig:ablation-ema-all}
    \vspace{-5pt}
\end{figure}

In this section, we implement extensive experiments to validate the effectiveness of each component, including three aspects for sample selection in Table~\ref{tab:ablation-three-factors}, the adaptive mechanism in Fig.~\ref{fig:ablation-adaptive} and model EMA in Fig.~\ref{fig:ablation-ema-all}.

\vspace{-5pt}

\vspace{-13pt}
\paragraph{The effect of three factors: Novelty \& Informativeness \& Diversity.}
In Table~\ref{tab:ablation-three-factors}, (a) denotes \texttt{Random}, (b) is random selection within new classes according to predictions, (c) denotes sampling informative instances from the entirety of samples predicted as new classes, and (d) is our method. The difference between (c) and (d) is that (d) samples informative instances in a class-wise manner, with $\lfloor b/K_{new}\rfloor$ of each new class to ensure diversity. (b) outperforms (a) by 2.05\% and 0.84\% on two datasets. (c) is slightly better than (b) for considering informativeness. Compared with (c), our method obtains consistent improvements in both old and new classes (+1.21\% and +3.47\% in old and new on CUB), indicating the importance of diversity.

\vspace{-5pt}
\paragraph{The effect of adaptive sampling.}
Our adaptive mechanism selects \emph{confident novel samples} at early rounds while \emph{informative samples} at later rounds. We compare our results with two singular baseline strategies in Fig.~\ref{fig:ablation-adaptive}. Ours consistently outperforms the other two strategies. For baseline methods, informative sampling outperforms confident sampling after five rounds. However, as shown in Fig.~\ref{fig:ablation-adaptive} (b), confident sampling works better at early rounds, which aligns with our rationale in Sec.~\ref{subsec:adaptive-novel-sampling}. That is, models require more confident samples for early learning of stable clusters.

\vspace{-5pt}
\paragraph{The effect of model EMA.}
As in Fig.~\ref{fig:ablation-ema-all} (b), we compute the consistency of \emph{label mapping function} computed on limited $\mathcal{D}_l^t$ and the whole test data. Results validate that EMA provides a more stable and consistent $\mathcal{M}$, which is suitable for \emph{inductive} evaluation, and obtains better results on various datasets in Fig.~\ref{fig:ablation-ema-all} (a).

\subsection{Further Analysis}

\begin{figure}[!t]
    \centering
    \includegraphics[width=0.9\linewidth]{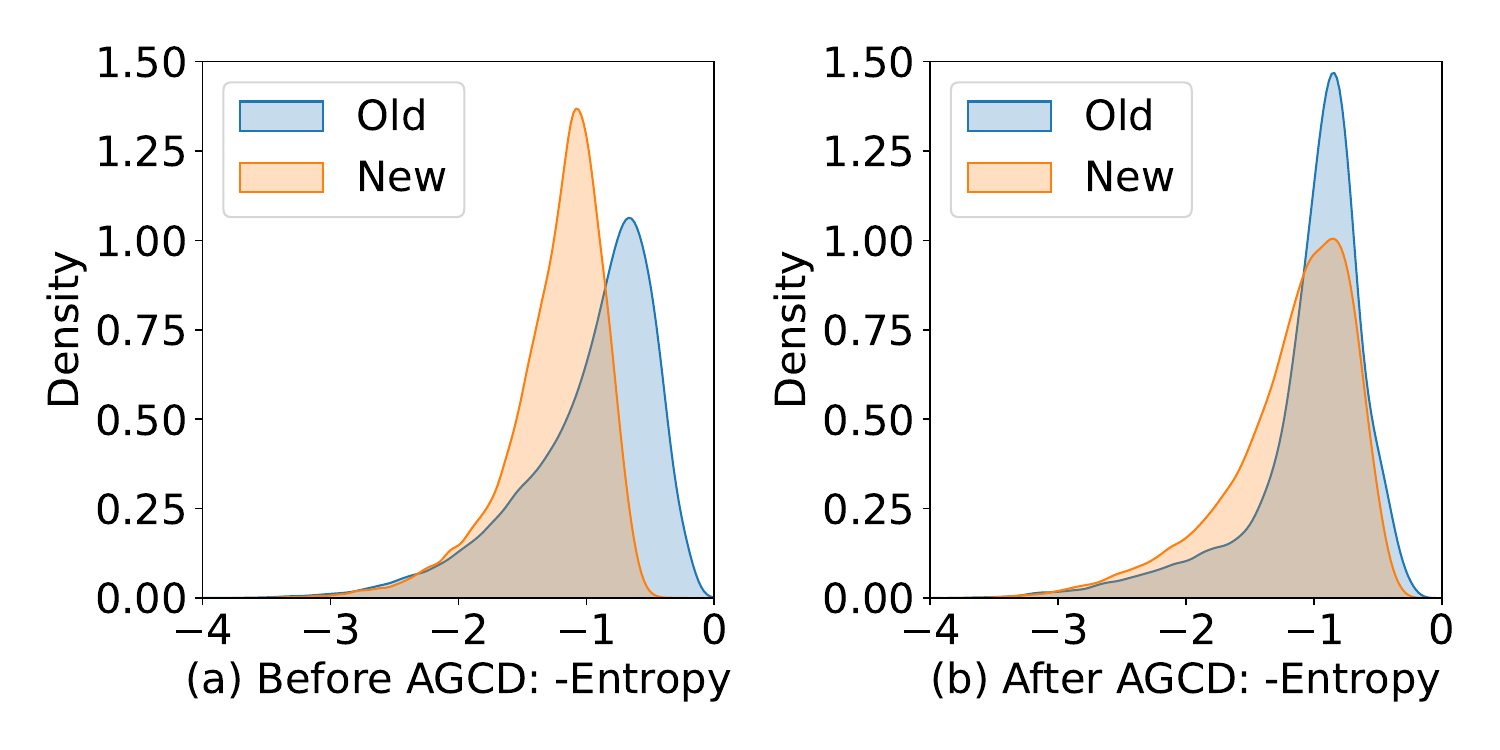}
    \vspace{-7pt}
    \caption{Confidence distribution before and after AGCD.}
    \label{fig:confidence-before-after-agcd}
    \vspace{-10pt}
\end{figure}

\paragraph{Confidence consistency.}
Fig.~\ref{fig:confidence-before-after-agcd} reveals that AGCD improves confidence consistency between old and new classes. Before AGCD, there was a noticeable gap between the confidence distributions of new and old classes, with a peak difference of $\sim 0.5$. After AGCD, the peak gap is almost reduced to zero. As a result, our method effectively addresses two issues of GCD including imbalanced accuracy and confidence with an affordable annotation budget.

\begin{table}[!ht]
\setlength\tabcolsep{5pt}
\centering
\renewcommand{\arraystretch}{1}
\caption{Results of All Acc on CUB of various initial label ratios.}
\vspace{-8pt}
\label{tab:cub-init-label-ratio}
\resizebox{.85\linewidth}{!}{
\begin{tabular}{@{}ccccccc@{}}
    \toprule
    label ratio & 0 & 0.01 & 0.05 & 0.1 & 0.2 & 0.3 \\ \midrule
    w/o AGCD & 14.15 & 18.12 & 27.68 & 37.49 & 50.17 & 58.49 \\
    \texttt{Random} & 31.46 & 32.00 & 45.88 & 53.16 & 62.74 & 66.45 \\
    \texttt{Entropy} & 32.02 & 36.95 & 46.32 & 53.78 & 62.82 & 60.55 \\
    \RC{30}Ours & \textbf{33.36} & \textbf{38.73} & \textbf{46.88} & \textbf{55.30} & \textbf{66.62} & \textbf{69.47} \\ \bottomrule
\end{tabular}
}
\vspace{-5pt}
\end{table}

\vspace{-5pt}
\begin{table}[!t]
\setlength\tabcolsep{5pt}
\centering
\renewcommand{\arraystretch}{1}
\caption{Results on CUB with various budge sizes per round $b$. Models are trained with 3 AGCD rounds.}
\vspace{-8pt}
\label{tab:various-num-query}
\resizebox{.8\linewidth}{!}{
\begin{tabular}{@{}cccccc@{}}
    \toprule
    $b$ & 30 & 50 & 100 & 300 & 500 \\ \midrule
    \texttt{Random} & 52.92 & 56.44 & 58.54 & 66.78 & 72.68 \\
    \texttt{Entropy} & 52.90 & 54.47 & 58.87 & 68.69 & 70.87 \\
    \RC{30}Ours & \textbf{53.40} & \textbf{56.50} & \textbf{59.86} & \textbf{69.59} & \textbf{73.56} \\ \bottomrule
\end{tabular}
}
\end{table}

\paragraph{Various label ratios and budget sizes.}
We conduct experiments with different settings, including various initial labeling ratios in Table~\ref{tab:cub-init-label-ratio} and various budget sizes $b$ in Table~\ref{tab:various-num-query}. The proposed strategy \texttt{Adaptive-Novel} consistently outperforms others across various settings, which showcases the sample selection aspects are general and robust to different settings in AGCD.

\begin{table}[!t]
\setlength\tabcolsep{5pt}
\centering
\renewcommand{\arraystretch}{1}
\caption{Results of AGCD with an unknown class number (estimated class number) on ImageNet-100 and CUB.}
\vspace{-8pt}
\label{tab:unknown-number}
\resizebox{.85\linewidth}{!}{
\begin{tabular}{@{}ccccccc@{}}
    \toprule
    \multirow{2}{*}{Strategies} & \multicolumn{3}{c}{CUB} & \multicolumn{3}{c}{ImageNet-100} \\ \cmidrule(l){2-4} \cmidrule(l){5-7} 
     & All & Old & New & All & Old & New \\ \midrule
    \texttt{Random} & 60.68 & 63.31 & 58.08 & 77.86 & \textbf{90.34} & 65.38 \\
    \texttt{Entropy} & 60.48 & 64.84 & 56.15 & 76.12 & 71.52 & 60.72 \\
    \RC{30}Ours & \textbf{64.14} & \textbf{66.09} & \textbf{62.20} & \textbf{82.46} & 89.84 & \textbf{70.64} \\ \bottomrule
\end{tabular}
}
\vspace{-10pt}
\end{table}

\vspace{-10pt}
\paragraph{Unknown class number.}
We also consider the scenarios with unknown class number $K_{new}$ in Table.~\ref{tab:unknown-number}. We perform an off-the-shelf number estimation algorithm~\cite{vaze2022generalized} to get an estimation of $K_{new}$ in advance, and use it to construct classifiers. Results in Table~\ref{tab:unknown-number} show that \texttt{Adaptive-Novel} is robust to the unknown class numbers, indicating the superiority of our method. Details are shown in the appendix.

\section{Conclusions}
\label{sec:conclusion}

In this paper, we propose a new setting of Active Generalized Category Discovery (AGCD) to address the inherent and intractable issues of GCD. Moreover, we pose two unique challenges in AGCD, \ie, new classes in unlabeled data and the clustering nature of GCD. To solve these challenges, we propose an adaptive query strategy \texttt{Adaptive-Novel} considering novelty, informativeness and diversity, which adaptively selects samples with proper uncertainty. Besides, we further propose a stable \emph{label mapping} algorithm to address the issue of different ordering of label indices between ground truth labels and models' label space. Experiments show that our method achieves state-of-the-art performance across different scenarios.


\vspace{10pt}
\noindent
\textbf{Acknowledgements} This work has been supported by the National Science and Technology Major Project (2022ZD0116500),  National Natural Science Foundation of China (U20A20223, 62222609, 62076236), CAS Project for Young Scientists in Basic Research (YSBR-083), and Key Research Program of Frontier Sciences of CAS (ZDBS-LY-7004).

\clearpage
\newpage
{
    \small
    \bibliographystyle{ieeenat_fullname}
    \bibliography{main}
}

\appendix

\maketitlesupplementary

\section*{Overview}
In this appendix, we provide additional descriptions of the following contents:
\begin{itemize}
    \item Fundamental principles of AGCD are discussed in Appendix~\ref{sec:appendix-principle}, including the relationship to some similar settings and the main concern of this paper.
    \item More details of the proposed sampling strategy are elaborated in Appendix~\ref{sec:appendix-adaptive-novel}.
    \item We provide more experimental details in Appendix~\ref{sec:appendix-implementaion}.
    \item Additional quantitative and qualitative results are provided in Appendix~\ref{sec:appendix-quantitative-results} and Appendix~\ref{sec:appendix-qualitative-results}, respectively.
\end{itemize}

\section{Fundamental Principles of AGCD}
\label{sec:appendix-principle}

In this section, we compare several related settings, and distinguish between sample selection strategies and training methods, to clarify the positioning of our contribution.

\vspace{-13pt}
\paragraph{Comparisons to related settings.}
We compare conventional Active Learning (AL), Generalized Category Discovery (GCD) and the proposed Active Generalized Category Discovery (AGCD):
\begin{itemize}
    \item AL aims to greatly improve models' performance with affordable labeling budgets. The spirit of AL is to select valuable and informative samples for labeling and incorporate newly annotated samples into the training set for subsequent training, which helps disambiguate confusing samples and correct previous errors. AL principally follows a closed-world setting, where unlabeled data contains the same set of classes as the labeled data.
    \item GCD is an open-world task, which aims to classify old classes and discover novel classes in the unlabeled data, given some labeled samples from old classes. However, GCD itself is not a fully learnable task in that the knowledge from old classes is not fully transferable to new classes and new classes are purely unlabeled, resulting in intractable problems including imbalanced accuracy and confidence between old and new classes, as in Sec.~\ref{subsec:gcd-problems}.
    \item AGCD is proposed to address the intractable problems and largely enhance the performance of GCD with minimal annotation costs. AGCD adopts a similar spirit to AL. Additionally, AGCD is a more general setting and generalizes AL to the open-world setting. We could refer to GCD as an extrapolated version of AL, where new classes could exist in unlabeled data, and models are required to classify both old and new classes. With additional labeled data, especially those from new classes, models could rectify previously inevitable errors and biases, and clarify the decision boundaries. Unlike GCD, models in AGCD are evaluated in \emph{inductive settings} with disjoint test datasets.
\end{itemize}

\paragraph{Sample selection strategies and training methods.}
AGCD mainly involves two components, including sample selection strategies and training methods. They are complementary to each other. Models first select informative samples to obtain annotations from the oracle, and are subsequently trained on them using specific training methods.

\paragraph{Main concern and focus of this paper.}
In this paper, our main concern is how to select valuable samples from the data perspective and why conventional AL strategies are not applicable to the task of AGCD, and we propose a sample selection strategy called \texttt{Adaptive-Novel}. To validate its effectiveness, we compare \texttt{Adaptive-Novel} with various strategies in the literature of AL, as elaborated in Appendix~\ref{sec:appendix-adaptive-novel}. For training methods, we directly employ the off-the-shelf method SimGCD~\cite{Wen_2023_ICCV} for all strategies for fair comparison, considering that SimGCD~\cite{Wen_2023_ICCV} is the SOTA method in GCD and it employs parametric classifiers which are efficient for training and convenient for evaluation including accuracy and confidence. By contrast, non-parametric methods~\cite{vaze2022generalized,pu2023dynamic} require independent K-Means~\cite{macqueen1967some} clustering, which is inefficient and non-trivial to analyze confidence and evaluate in \emph{inductive} settings.

Although the training methodology, \ie, the loss function, is not the focus of this paper, we still need to address specific issues over the course of AGCD training, \ie, different ordering of label indices in clustering problems. Specifically, in GCD, we implement clustering on new classes, the core is to cluster samples of the same novel class together and separate samples from different classes. The specific assignment of the labels to each cluster is not crucial. As a result, the label indices of new classes differ across various experiments and are unpredictable. That is the reason why it is common to employ the Hungarian algorithm~\cite{kuhn1955hungarian} to evaluate the performance of GCD, as in Eq.~\eqref{eq:gcd-acc-evaluation}. For example, the model might assign ``8'' and ``9'' to ``birds'' in two runs. This brings about a challenging issue in AGCD, \ie, the queried ground truth labels could not be directly used by the model due to the different ordering between the model's predictions and ground truth labels. As a result, it is necessary to obtain a \emph{mapping function} to convert the ground truth labels to the label space of the classifier in advance, as discussed in Sec.~\ref{subsec:stable-label-mapping}. Considering that the labeled data $\mathcal{D}_l^t$ is limited, especially for new classes, this would introduce instability to the computation of \emph{label mapping} $\mathcal{M}^t$. To overcome this issue, we propose a stable labeling mapping method. Specifically, we perform Hungarian optimal assignment algorithm~\cite{kuhn1955hungarian} between the EMA model's predictions $\hat y_i^{ema}$ and the ground truth labels $y_i$ over the accessible labeled data of the current round, \ie, $\mathcal{D}_l^t=\mathcal{D}_l^{t-1}\cup\mathcal{D}_q^t$, as in Eq.~\eqref{eq:label-mapping}. The EMA model could offer more stable predictions during training. Additionally, \emph{label mapping} $\mathcal{M}^t$ is computed in every iteration, so do the mapped ground truth labels $\mathcal{M}(y_i)$ and $\mathcal{D}_{l,map}^{t-1}\cup\mathcal{D}_{q,map}^t$, because the model's parameters are continuously updated, especially at the beginning of each round, we should also keep the \emph{label mapping} up-to-date to fit the changing model.

\section{More Details about \texttt{Adaptive-Novel}}
\label{sec:appendix-adaptive-novel}

\paragraph{Adaptive sampling mechanism.}
In \texttt{Adaptive-Novel} strategy, we propose an adaptive mechanism, \ie, at early rounds, we sample \emph{confident novel samples} to stabilize new clusters while \emph{informative novel samples} at later rounds to refine decision boundaries. We propose to transfer from the former to the latter when the clusters of new classes are stable. Technically, we compare the \emph{label mapping} at the initial $\mathcal{M}^t_{init}$ and final epochs $\mathcal{M}^t_{final}$ of current round $t$, when the difference is lower than a pre-defined threshold $\delta$, the clusters are deemed stable, and we transfer to seek for \emph{informative novel samples} at the next round $t+1$, the difference is computed as follows:
\begin{equation}
    \text{diff} = \frac{\sum_{i=1}^K \mathds{1}(\mathcal{M}^t_{init}[i]\neq \mathcal{M}^t_{final}[i])}{K}
    \label{eq:appendix-diff-mapping}
\end{equation}
where $K=K_{old}+K_{new}$ denotes the total number of classes, and $\mathcal{M}^t[i]$ denotes the mapping $\mathcal{M}^t$ from the $i$-th label index of the ground truth labels to the $\mathcal{M}^t[i]$-th classifier's predictive label index. In all our experiments, we set $\delta=0.1$ for all six datasets.

\vspace{-10pt}
\paragraph{Uncertainty metrics.}
We mainly consider three uncertainty/confidence metrics, including maximum softmax probability (MSP~\cite{hendrycks2017a}) $p(\hat y_1|\mathbf{x})$, margin $p(\hat y_1|\mathbf{x})-p(\hat y_2|\mathbf{x})$ and entropy $-\sum_{i=1}^K p(y=i|\mathbf{x})\log p(y=i|\mathbf{x})$, where $\hat y_1=\argmax_{y}p(y|\mathbf{x})$ and $\hat y_2=\argmax_{y\neq \hat y_1}p(y|\mathbf{x})$ represent two most likely labels of sample $\mathbf{x}$. Informative samples refer to those with maximum uncertainty, \ie, minimum MSP value, minimum margin value and maximum entropy value. In the setting AGCD, we found that margin is more robust and could select more novel samples, as a result, we choose margin as the uncertainty metric in \texttt{Adaptive-Novel}. We further provide results of different metrics in Appendix~\ref{sec:appendix-quantitative-results}.

\vspace{-10pt}
\paragraph{\texttt{Adaptive-Novel} sampling algorithm.}
Here we give the exact algorithm of \texttt{Adaptive-Novel} as in Algorithm~\ref{alg:adaptive-novel}. We highlight three aspects for sample selection, including novelty, informativeness and diversity, with red colors. Note that the \emph{label mapping} is performed in each iteration when the model is updated.

\begin{algorithm*}[p]
    \small
    \caption{\texttt{Adaptive-Novel} Sampling Strategy for AGCD}
    \label{alg:adaptive-novel}
    \begin{algorithmic}[1]
        \Require {Initial labeled dataset $\mathcal{D}_l^0$ and unlabeled dataset $\mathcal{D}_u^0$.}
        \Require {Total rounds $N$ and labeling budget per round $b$.}
        \Require {Total class number $K=K_{old}+K_{new}$ (The ground truth or estimated).}
        \Require {Stable mapping threshold $\delta$, initial transfer scalar $\mathcal{T}=\texttt{False}$.}
        \Require {Total epochs $E$ of each round.}
        \Require {EMA decay parameter $\beta$.}
        \State {Initialize $\mathcal{D}_l^0$ and $\mathcal{D}_u^0$ as in Table~\ref{tab:datasets-settings}.}
        \For {current round $t=1\to N$}
            \For {$c=1\to K_{new}$} \hfill\texttt{\# Class-wise sampling for \textcolor{red}{Diversity}}
                \If {$\mathcal{T}$} \hfill\texttt{\# Adaptive \textcolor{red}{Informativeness}: informative sampling at later rounds}
                \State $\triangleright$ Select $\lfloor b/K_{new}\rfloor$ samples with \textbf{minimum} margin from the $c$-th predictive novel class ($\hat y=K_{old}+c$) \hfill\texttt{\# \textcolor{red}{Novelty}} \\
                \qquad\qquad\quad~~ and query their labels to obtain $\mathcal{D}_{q,c}^t$
                \Else \hfill\texttt{\# Adaptive \textcolor{red}{Informativeness}: confident sampling at early rounds}
                \State $\triangleright$ Select $\lfloor b/K_{new}\rfloor$ samples with \textbf{maximum} margin from the $c$-th predictive novel class ($\hat y=K_{old}+c$) \hfill\texttt{\# \textcolor{red}{Novelty}} \\
                \qquad\qquad\quad~~ and query their labels to obtain $\mathcal{D}_{q,c}^t$
                \EndIf
            \EndFor
            \State $\triangleright$ All the queried data of the current round:
            \State $\quad\quad \mathcal{D}_q^t=\mathcal{D}_{q,1}^t\cup\mathcal{D}_{q,2}^t\cdots\cup \mathcal{D}_{q,K_{new}}^t$
            \State $\triangleright$ Update labeled and unlabeled datasets:
            \State $\quad\quad \mathcal{D}_l^t=\mathcal{D}_l^{t-1}\cup\mathcal{D}_q^t$
            \State $\quad\quad \mathcal{D}_u^t=\mathcal{D}_u^{t-1}\setminus\mathcal{D}_q^t$
            \For {current epoch $e=1\to E$} \hfill\texttt{\# Training at the current round}
                \State $\triangleright$ Obtain \emph{label mapping function} $\mathcal{M}^t$ between ground truth labels $y_i$ and the EMA model predictions $\hat y_i^{ema}$ in Eq.~\eqref{eq:label-mapping} on $\mathcal{D}_l^t$
                \State $\triangleright$ Perform \emph{label mapping} on $\mathcal{D}_l^t$:
                \State $\quad\quad \mathcal{D}_{l,map}^t=\mathcal{D}_{l,map}^{t-1}\cup\mathcal{D}_{q,map}^t=\mathcal{M}^t(\mathcal{D}_l^{t-1})\cup \mathcal{M}^t(\mathcal{D}_q^t)$
                \State $\triangleright$ Train the model on $\mathcal{D}_{l,map}^t$ with supervised loss $\mathcal{L}_{con}^l$ in Eq.~\eqref{eq:loss-con-l} and $\mathcal{L}_{cls}^l$ in Eq.~\eqref{eq:loss-simgcd-cls-l} and unsupervised loss, \\
                \qquad\qquad~ and on $\mathcal{D}_u^t$ with purely unsupervised loss
                \State $\triangleright$ Update the EMA model with decay rate $\beta$
            \EndFor
            \State $\triangleright$ Compute the difference between $\mathcal{M}^t_{init}$ and $\mathcal{M}^t_{final}$ of this round as in Eq.~\eqref{eq:appendix-diff-mapping}
            \If {$\text{diff}<\delta$} \hfill\texttt{\# Mapping is stable and we transfer to informative sampling from round $t+1$}
            \State $\mathcal{T}=\texttt{True}$
            \EndIf
        \EndFor
        \Ensure {The trained model and datasets $\mathcal{D}_l^N$, $\mathcal{D}_u^N$ after $N$ AGCD rounds.}
    \end{algorithmic}
\end{algorithm*}

\section{More Experimental Details}
\label{sec:appendix-implementaion}

\paragraph{Comparative strategies.}
We compare the proposed \texttt{Adaptive-Novel} with various query strategies in the literature of conventional AL, including uncertainty-based and diversity/representative-based methods:
\begin{itemize}
    \item \texttt{Random}: a baseline that randomly selects samples from the unlabeled dataset.
    \item \texttt{Entropy}~\cite{wang2014new}: an uncertainty-based method that selects samples with the highest entropy over all classes $-\sum_{i=1}^K p(y=i|\mathbf{x})\log p(y=i|\mathbf{x})$.
    \item \texttt{LeastConf}~\cite{wang2014new}: an uncertainty-based method that selects samples with the lowest MSP~\cite{hendrycks2017a} $p(\hat y_1|\mathbf{x})$.
    \item \texttt{Margin}~\cite{roth2006margin}: an uncertainty-based method that selects samples with the lowest margin $p(\hat y_1|\mathbf{x})-p(\hat y_2|\mathbf{x})$.
    \item \texttt{KMeans}~\cite{macqueen1967some}: a diversity-based method that selects samples closest to the centroids of K-Means, which is implemented in the embedding space in a cluster-wise manner.
    \item \texttt{CoreSet}~\cite{sener2018active} picks up unlabeled samples with the greatest distances to their nearest labeled neighbor, and obtains representative samples of unlabeled data.
    \item \texttt{BADGE}~\cite{Ash2020Deep} is short for Batch Active learning by Diverse Gradient Embeddings and could be viewed as a hybrid method to query centroids from K-Means clustering over the gradient embeddings.
\end{itemize}

\vspace{-7pt}
\paragraph{Other implementation details.}
We adopt all training parameters from SimGCD~\cite{Wen_2023_ICCV}. The weight of supervised loss $\lambda$ is 0.35, and the weight $\lambda_e$ depends on specific datasets, we set $\lambda_e=1$ for CIFAR10, Aircraft and Stanford Cars, and 2 for ImageNet-100 and CUB, while 4 for CIFAR100. The temperature $\tau_c,\tau_p$ are 0.07 and 0.1 respectively. The sharpened temperature in self-distillation in Eq.~\eqref{eq:loss-simgcd-cls-u} is the same as SimGCD~\cite{Wen_2023_ICCV}, \ie, ramp-up schedule from 0.07 to 0.04. As for the hyper-parameters related to \texttt{Adaptive-Novel}, the EMA decay rate $\beta=0.9$, and we set the threshold $\delta$ for justification stability of \emph{label mapping function} to be 0.1 for all datasets. We first train models with SimGCD for 200 epochs as the base training stage to initialize models for subsequent AGCD. At each round of AGCD, we train models for 15 epochs. The default setting is to query 100 samples per round, and five rounds in total, CIFAR10 is an exception with only one round. During AGCD, we separately train all the queried data till the current round $\mathcal{D}_{q,map}^{all,t}=\mathcal{D}_{q,map}^1\cup\mathcal{D}_{q,map}^2\cup\cdots\cup \mathcal{D}_{q,map}^t$ and the original data $\mathcal{D}_l^0,\mathcal{D}_u^0$. We choose a smaller batch size 8 for $\mathcal{D}_{q,map}^{all,t}$ to acquire more update iterations and keep batch size 128 for the original dataset. The parameters above are applied to all query strategies for fair comparisons. All labeled data including $\mathcal{D}_l^0$ and $\mathcal{D}_{q,map}^{all,t}$ are trained with supervised objectives $\mathcal{L}_{con}^l$ Eq.~\eqref{eq:loss-con-l} and $\mathcal{L}_{cls}^l$ in Eq.~\eqref{eq:loss-simgcd-cls-l}. We also employ unsupervised loss $\mathcal{L}_{con}^u$ in Eq.~\eqref{eq:loss-con-u} and $\mathcal{L}_{cls}^u$ in Eq.~\eqref{eq:loss-simgcd-cls-u} on both labeled and unlabeled data, which is consistent with SimGCD.

\clearpage
\section{Additional Quantitative Results}
\label{sec:appendix-quantitative-results}

In this section, we provide additional quantitative results beyond the main text.

\begin{figure}[!ht]
    \centering
    \begin{subfigure}[b]{0.48\linewidth}
        \centering
        \includegraphics[width=\linewidth]{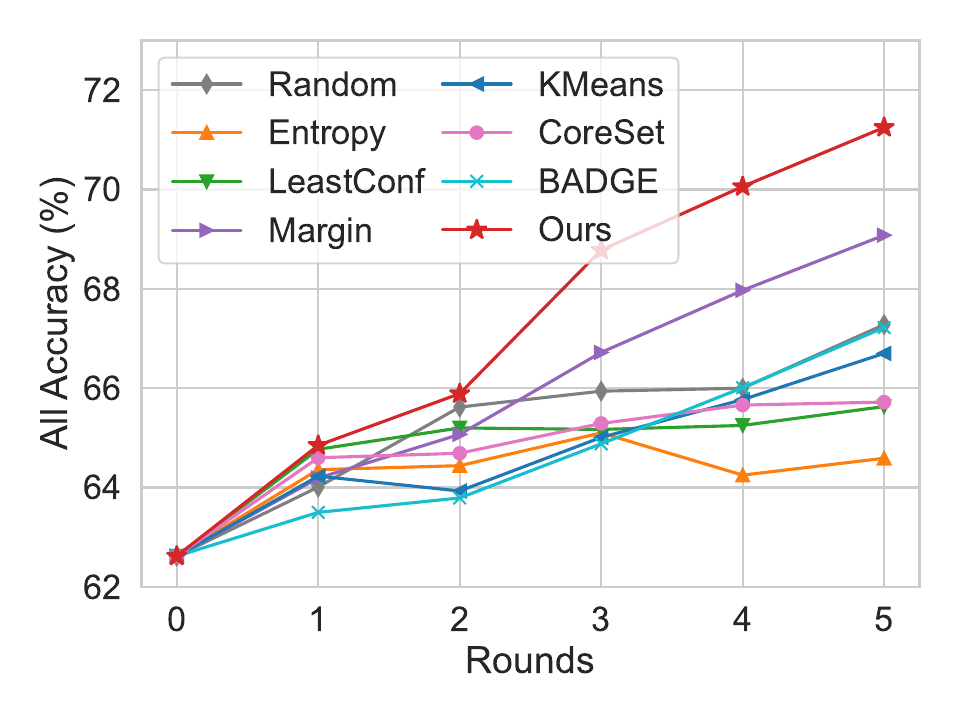}
        \caption{CIFAR100.}
        \label{subfig:appendix-5-rounds-cifar100}
    \end{subfigure}
    \hfill
    \begin{subfigure}[b]{0.48\linewidth}
        \centering
        \includegraphics[width=\textwidth]{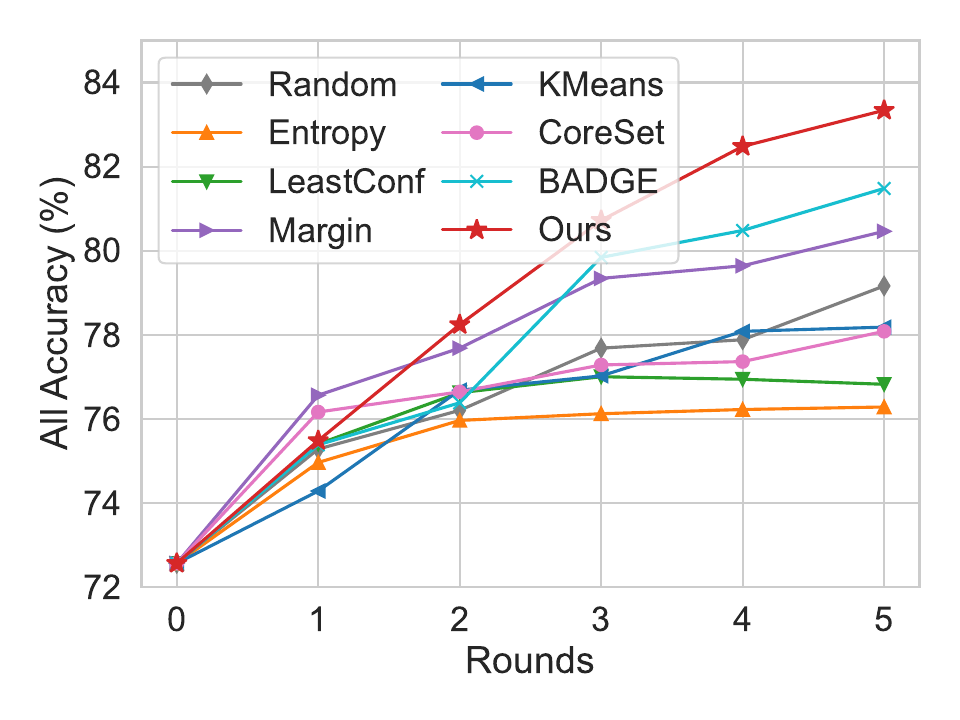}
        \caption{ImageNet-100.}
        \label{subfig:appendix-5-rouinds-imagenet100}
    \end{subfigure}
    \vspace{15pt}
    \begin{subfigure}[b]{0.48\linewidth}
        \centering
        \includegraphics[width=\linewidth]{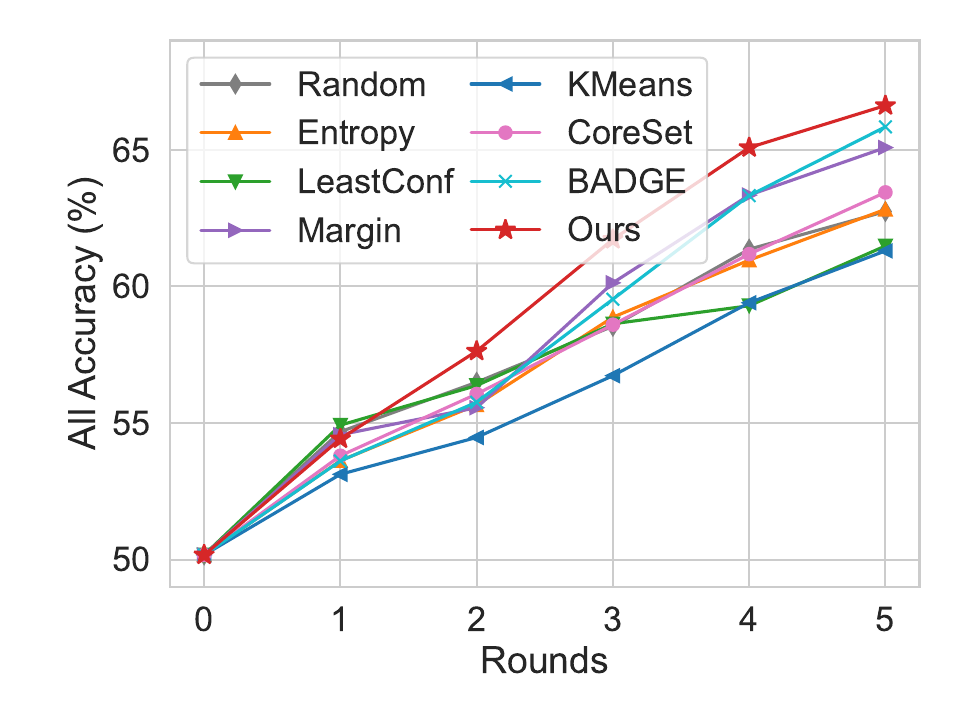}
        \caption{CUB.}
        \label{subfig:appendix-5-rounds-cub}
    \end{subfigure}
    \hfill
    \begin{subfigure}[b]{0.48\linewidth}
        \centering
        \includegraphics[width=\textwidth]{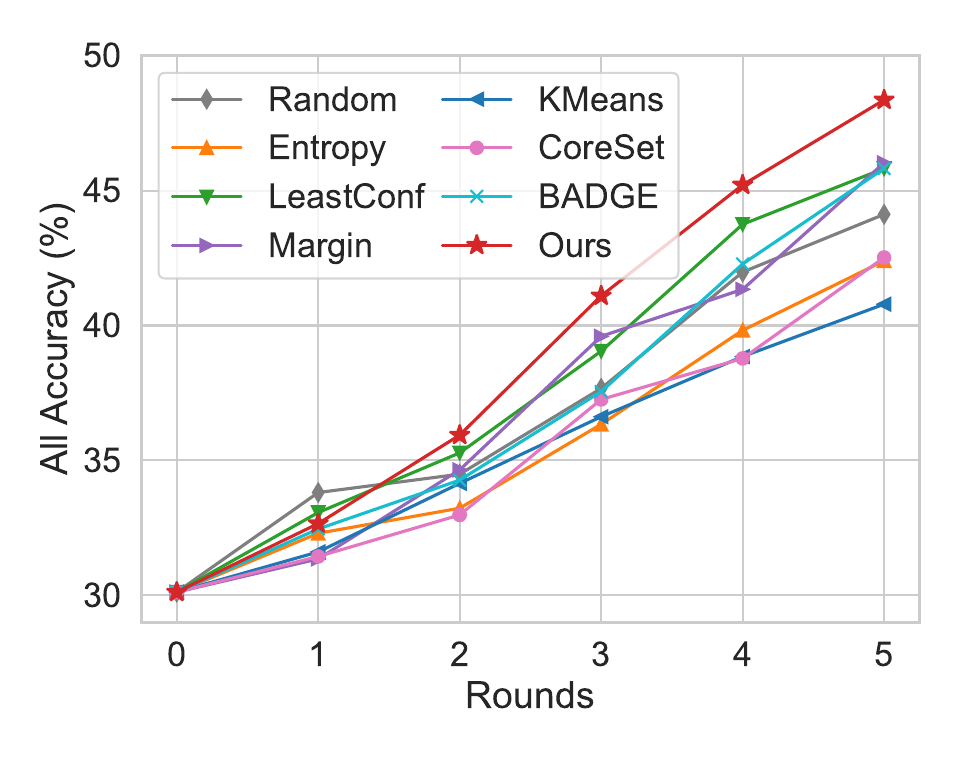}
        \caption{Stanford Cars.}
        \label{subfig:appendix-5-rouinds-scars}
    \end{subfigure}
    \\
    \begin{subfigure}[b]{0.48\linewidth}
        \centering
        \vspace{-10pt}
        \includegraphics[width=\textwidth]{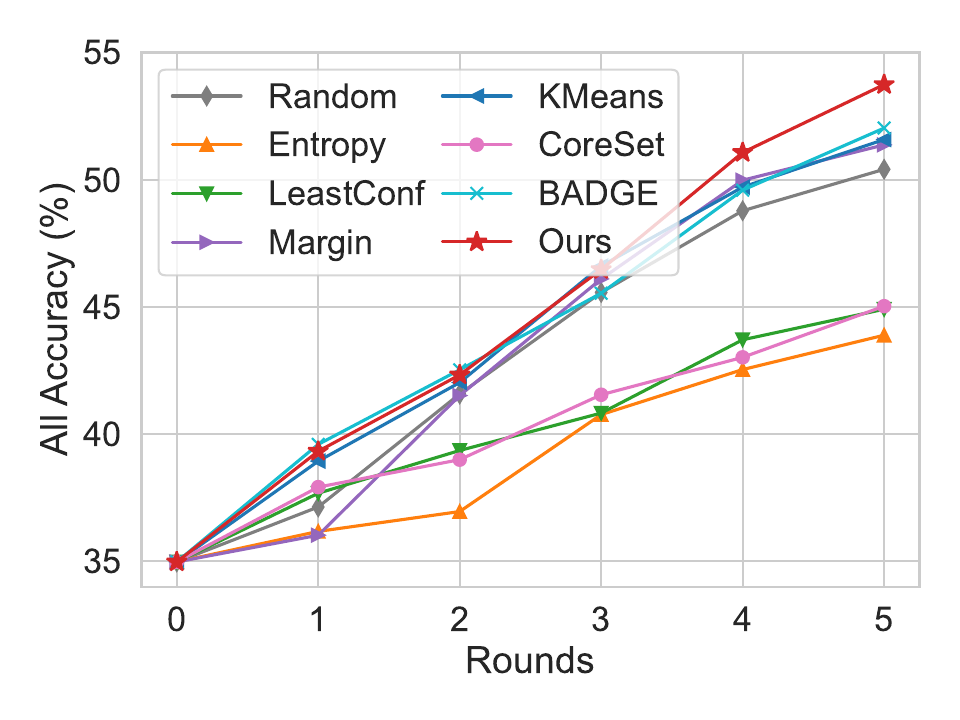}
        \caption{Aircraft.}
        \label{subfig:appendix-5-rouinds-aircraft}
    \end{subfigure}
    \vspace{-5pt}
    \caption{All accuracy of different strategies over 5 rounds.}
    \label{fig:appendix-5-rounds}
\end{figure}

\vspace{-10pt}
\paragraph{Details of AGCD performance across five rounds.}
In the main text, we mainly report the performance after all rounds of AGCD in Table~\ref{tab:main-acc-generic} and Table~\ref{tab:main-acc-fine}. Here, we provide more detailed results performance over the course of different AGCD rounds, as shown in Fig.~\ref{fig:appendix-5-rounds}, where mean results over three runs are plotted.


\vspace{-10pt}
\paragraph{Results of transfer rounds.}
As discussed in Sec.~\ref{subsec:adaptive-novel-sampling} and Appendix~\ref{sec:appendix-adaptive-novel}, when \emph{label mapping function} is stable at round $t$, \ie, $\text{diff}<\delta$, we transfer from \emph{confident novel sampling} to \emph{informative novel sampling} from round $t+1$. Results of the transfer round $t+1$ are shown in Table~\ref{tab:appendix-transfer-round}, including five datasets CIFAR100 (C-100), ImageNet-100 (IN-100), CUB, Stanford Cars (SCars) and FGVC-Aircraft (Air). As for CIFAR10, the default setting has only one round, and we adopt \emph{informative novel sampling} on it.

\begin{table}[!htb]
\scriptsize
\setlength\tabcolsep{6pt}
\centering
\renewcommand{\arraystretch}{1}
\caption{Transfer round $t+1$ from \emph{confident novel sampling} to \emph{informative novel sampling} on various datasets.}
\vspace{-7pt}
\label{tab:appendix-transfer-round}
\resizebox{.98\linewidth}{!}{
\begin{tabular}{@{}cccccc@{}}
\toprule
Datasets & C-100 & IN-100 & CUB & SCars & Air \\ \midrule
Round $t+1$ & 2 & 2 & 4 & 2 & 2 \\ \bottomrule
\end{tabular}
}
\end{table}

\vspace{-10pt}
\paragraph{\texttt{Adaptive-Novel} with different uncertainty metrics.}
As introduced in Sec.~\ref{subsec:adaptive-novel-sampling}, we choose margin as the metric for uncertainty/confidence for \texttt{Adaptive-Novel}, owing to the fact that margin is more robust in the task of AGCD and could select more sample from new classes when compared with MSP and entropy. From Fig.~\ref{fig:old-new-confidence} we can observe that there are more samples from novel categories in the low confidence regime of Margin compared with Entropy and MSP. The results in Table~\ref{tab:novelty-c100-cars} also show that we could select more new samples when using \texttt{Margin} as a metric for uncertainty-based methods. Here, we also conduct experiments using our \texttt{Adaptive-Novel} strategy but with different uncertainty metrics, as shown in Table~\ref{tab:appendix-uncertainty-metric}. Results indicate the superiority of Margin benefiting from selecting more samples from novel categories.

\begin{table}[!h]
\setlength\tabcolsep{6pt}
\centering
\renewcommand{\arraystretch}{1}
\caption{Results of \texttt{Adaptive-Novel} when applying different uncertainty metrics, including Entropy, MSP and Margin.}
\vspace{-7pt}
\label{tab:appendix-uncertainty-metric}
\resizebox{.98\linewidth}{!}{
\begin{tabular}{@{}ccccccc@{}}
\toprule
\multirow{2}{*}{Metrics} & \multicolumn{3}{c}{CIFAR100} & \multicolumn{3}{c}{CUB} \\ \cmidrule(l){2-4} \cmidrule(l){5-7}
 & All & Old & New & All & Old & New \\ \midrule
Ours w/ Entropy & 67.30 & \textbf{76.94} & 57.66 & 63.75 & \textbf{69.22} & 58.34 \\
Ours w/ MSP & 67.63 & 75.18 & 60.08 & 63.70 & 67.37 & 60.07 \\
Ours w/ Margin & \textbf{71.25} & 75.72 & \textbf{66.78} & \textbf{66.62} & 66.54 & \textbf{66.70} \\ \bottomrule
\end{tabular}
}
\end{table}

\vspace{-10pt}
\paragraph{Unknown class number scenarios.}
SimGCD~\cite{Wen_2023_ICCV} is a parametric classifier, it requires the number of classes $K$ or $K_{new}$ is known \emph{a-prior} before training, here it could be the ground truth or estimated with off-the-shelf methods. In this paper, we adopt the off-the-shelf method \texttt{Max-ACC}~\cite{vaze2022generalized} to estimate the number of the new classes. \texttt{Max-ACC} performs K-Means clustering on the entire dataset with various number of new classes, and choose the value as an estimation corresponding to the maximum clustering accuracy of labeled data. The estimation results on several datasets are shown in Table~\ref{tab:appendix-estimation-class-number}.
Then we directly use the estimated number $\widehat K$ to set the prototypical classifier $\mathcal{C}=\{\mathbf{c}_1,\mathbf{c}_2,\cdots,\mathbf{c}_{\widehat K}\}$. In our main text, we have provided results on ImageNet-100 and CUB as in Table~\ref{tab:unknown-number}. Here we show results on more datasets in Table~\ref{tab:appendix-unknown-number}.

\begin{table}[!htb]
\setlength\tabcolsep{4pt}
\centering
\renewcommand{\arraystretch}{1}
\caption{Estimation of total class number $\widehat K=K_{old}+\widehat K_{new}$ using \texttt{Max-ACC}~\cite{vaze2022generalized}.}
\vspace{-7pt}
\label{tab:appendix-estimation-class-number}
\resizebox{.9\linewidth}{!}{
\begin{tabular}{@{}ccccc@{}}
\toprule
Datasets & CIFAR100 & ImageNet-100 & CUB & Stanford Cars \\ \midrule
Ground Truth & 100 & 100 & 200 & 196 \\
Estimated & 100 & 109 & 231 & 230 \\ \bottomrule
\end{tabular}
}
\end{table}

\begin{table*}[!ht]
\setlength\tabcolsep{6pt}
\centering
\renewcommand{\arraystretch}{1}
\caption{Results of AGCD with unknown class number (estimated class number) on various datasets.}
\vspace{-7pt}
\label{tab:appendix-unknown-number}
\resizebox{.8\textwidth}{!}{
\begin{tabular}{@{}ccccccccccccc@{}}
\toprule
\multirow{2}{*}{Strategies} & \multicolumn{3}{c}{CIFAR100} & \multicolumn{3}{c}{ImageNet-100} & \multicolumn{3}{c}{CUB} & \multicolumn{3}{c}{Stanford Cars} \\ \cmidrule(l){2-4} \cmidrule(l){5-7} \cmidrule(l){8-10} \cmidrule(l){11-13} 
 & All & Old & New & All & Old & New & All & Old & New & All & Old & New \\ \midrule
\texttt{Random} & 67.28 & 74.52 & 60.04 & 77.86 & \textbf{90.34} & 65.38 & 60.68 & 63.31 & 58.08 & 42.22 & 55.24 & 29.66 \\
\texttt{Entropy} & 64.59 & 73.94 & 55.24 & 76.12 & 71.52 & 60.72 & 60.48 & 64.84 & 56.15 & 41.77 & 56.36 & 27.71 \\
\RC{30}Ours & \textbf{71.25} & \textbf{75.72} & \textbf{66.78} & \textbf{82.46} & 89.84 & \textbf{70.64} & \textbf{64.14} & \textbf{66.09} & \textbf{62.20} & \textbf{43.92} & \textbf{58.05} & \textbf{30.30} \\ \bottomrule
\end{tabular}
}
\end{table*}

\vspace{-10pt}
\paragraph{Results about novelty metrics.}
In the main text, results about novelty metrics on CIFAR100 and Stanford Cars are shown in Table~\ref{tab:novelty-c100-cars}. Here we provide the results on all six datasets as in Table~\ref{tab:appendix-novelty-generic} and Table~\ref{tab:appendix-novelty-fine}. Our method consistently selects more samples evenly distributed across novel categories, leading to better AGCD performance.

\begin{table*}[!t]
\setlength\tabcolsep{2pt}
\centering
\renewcommand{\arraystretch}{1}
\caption{Novelty metrics of selected data on generic datasets.}
\vspace{-7pt}
\label{tab:appendix-novelty-generic}
\resizebox{.95\linewidth}{!}{
\begin{tabular}{@{}ccccccccccccc@{}}
\toprule
\multirow{2}{*}{AL Strategies} & \multicolumn{4}{c}{CIFAR10} & \multicolumn{4}{c}{CIFAR100} & \multicolumn{4}{c}{ImageNet-100} \\ \cmidrule(l){2-5} \cmidrule(l){6-9} \cmidrule(l){10-13} 
 & $\texttt{Nov-C}$ & $\texttt{Nov-R}$ & $\texttt{Nov-U}$ & $\texttt{Nov-I}$ & $\texttt{Nov-C}$ & $\texttt{Nov-R}$ & $\texttt{Nov-U}$ & $\texttt{Nov-I}$ & $\texttt{Nov-C}$ & $\texttt{Nov-R}$ & $\texttt{Nov-U}$ & $\texttt{Nov-I}$ \\ \midrule
\texttt{Random} & \textbf{1.00} & 0.84 & 0.96 & 0.81 & \textbf{1.00} & 0.52 & 0.97 & 0.50 & \textbf{1.00} & 0.56 & 0.97 & 0.54 \\
\texttt{Entropy} & \textbf{1.00} & 0.62 & 0.90 & 0.56 & 0.90 & 0.44 & 0.91 & 0.40 & 0.88 & 0.40 & 0.90 & 0.36 \\
\texttt{Margin} & \textbf{1.00} & 0.89 & 0.90 & 0.80 & 0.96 & 0.63 & 0.95 & 0.60 & 0.98 & 0.78 & 0.95 & 0.74 \\
\texttt{CoreSet} & \textbf{1.00} & 0.83 & 0.96 & 0.80 & 0.96 & 0.61 & 0.94 & 0.57 & 0.98 & 0.56 & 0.96 & 0.54 \\
\texttt{BADGE} & 0.88 & 0.77 & 0.87 & 0.67 & \textbf{1.00} & 0.63 & \textbf{0.98} & 0.62 & \textbf{1.00} & 0.71 & 0.97 & 0.69 \\
\RC{30}Ours & \textbf{1.00} & \textbf{0.90} & \textbf{0.97} & \textbf{0.87} & \textbf{1.00} & \textbf{0.71} & \textbf{0.98} & \textbf{0.70} & \textbf{1.00} & \textbf{0.82} & \textbf{0.98} & \textbf{0.80} \\ \bottomrule
\end{tabular}
}
\end{table*}

\begin{table*}[!t]
\setlength\tabcolsep{2pt}
\centering
\renewcommand{\arraystretch}{1}
\caption{Novelty metrics of selected data on fine-grained datasets.}
\vspace{-7pt}
\label{tab:appendix-novelty-fine}
\resizebox{.95\linewidth}{!}{
\begin{tabular}{@{}ccccccccccccc@{}}
\toprule
\multirow{2}{*}{AL Strategies} & \multicolumn{4}{c}{CUB} & \multicolumn{4}{c}{Stanford Cars} & \multicolumn{4}{c}{FGVC-Aircraft} \\ \cmidrule(l){2-5} \cmidrule(l){6-9} \cmidrule(l){10-13} 
 & $\texttt{Nov-C}$ & $\texttt{Nov-R}$ & $\texttt{Nov-U}$ & $\texttt{Nov-I}$ & $\texttt{Nov-C}$ & $\texttt{Nov-R}$ & $\texttt{Nov-U}$ & $\texttt{Nov-I}$ & $\texttt{Nov-C}$ & $\texttt{Nov-R}$ & $\texttt{Nov-U}$ & $\texttt{Nov-I}$ \\ \midrule
\texttt{Random} & 0.95 & 0.63 & 0.96 & 0.60 & 0.93 & 0.57 & 0.96 & 0.55 & 0.96 & 0.53 & 0.98 & 0.52 \\
\texttt{Entropy} & 0.78 & 0.58 & 0.91 & 0.53 & 0.85 & 0.64 & 0.92 & 0.59 & 0.92 & 0.57 & 0.92 & 0.52 \\
\texttt{Margin} & 0.87 & 0.59 & 0.95 & 0.56 & 0.90 & 0.66 & 0.93 & 0.61 & \textbf{1.00} & 0.66 & 0.98 & 0.65 \\
\texttt{CoreSet} & 0.93 & 0.71 & 0.95 & 0.67 & 0.89 & \textbf{0.69} & 0.94 & 0.65 & 0.94 & 0.58 & 0.95 & 0.55 \\
\texttt{BADGE} & 0.98 & 0.57 & 0.97 & 0.55 & 0.95 & 0.64 & \textbf{0.97} & 0.62 & 1.00 & 0.59 & 0.98 & 0.58 \\
\RC{30}Ours & \textbf{1.00} & \textbf{0.75} & \textbf{0.98} & \textbf{0.74} & \textbf{0.98} & \textbf{0.69} & \textbf{0.97} & \textbf{0.67} & \textbf{1.00} & \textbf{0.76} & \textbf{0.99} & \textbf{0.75} \\ \bottomrule
\end{tabular}
}
\vspace{5pt}
\end{table*}

\paragraph{Performance on the long-tailed dataset.}
In the real world, the class distributions often follow a long-tailed distribution, where the head classes have significantly more samples than the tail classes. Several methods~\cite{zhang2023novel,NEURIPS2023_b7216f4a} have explored this issue in the task of GCD. In the long-tailed settings, the uniform constraint $H(\overline{\mathbf{p}})$ of the SimGCD~\cite{Wen_2023_ICCV} training procedure could be less applicable, as a result, we assign a small value to the weight $\lambda_e$ to balance between addressing imbalanced class distribution and avoiding trivial solutions. In this paper, we also test our methods in this realistic scenarios. Specifically, we conduct experiments on the long-tailed Herbarium19~\cite{tan2019herbarium} dataset. Table~\ref{tab:appendix-herb19} shows that our strategy outperforms other competitors, demonstrating the strong applicability of our method.

\begin{table}[!h]
    \setlength\tabcolsep{6pt}
    \centering
    \renewcommand{\arraystretch}{.8}
    \caption{Performance on the long-tailed Herbarium19 dataset.}
    \vspace{-7pt}
    \label{tab:appendix-herb19}
    \resizebox{.65\linewidth}{!}{
    \begin{tabular}{@{}cccc@{}}
    \toprule
    Strategies & All & Old & New \\ \midrule
    w/o AGCD & 46.55 & 62.67 & 29.49 \\
    \texttt{Entropy} & 52.44 & \textbf{65.36} & 38.79 \\
    \texttt{BADGE} & 53.31 & 61.53 & 44.62 \\ \midrule
    \RC{30}Ours & \textbf{54.50} & 63.33 & \textbf{45.16} \\ \bottomrule
    \end{tabular}
    }
\end{table}

\paragraph{Performance under other GCD training procedures.}
In the main manuscript, we compare different sample selection strategies with the GCD training procedure SimGCD~\cite{Wen_2023_ICCV}. Here we present results under two recent GCD training methods, \ie, $\mu$GCD~\cite{NEURIPS2023_3f52ab43} and PIM~\cite{chiaroni2023parametric}. As shown in Table~\ref{tab:appendix-two-training-backbone}, \texttt{Adaptive-Novel} works across various GCD training methods, indicating the superiority of our strategy consistency.

\begin{table}[!h]
    \setlength\tabcolsep{3pt}
    \centering
    \footnotesize
    \renewcommand{\arraystretch}{.8}
    \caption{Performance with another two GCD training procedures, $\mu$GCD and PIM.}
    \vspace{-7pt}
    \label{tab:appendix-two-training-backbone}
    \resizebox{.8\linewidth}{!}{
    \begin{tabular}{@{}ccccccc@{}}
    \toprule
    \multirow{2}{*}{Strategies} & \multicolumn{3}{c}{$\mu$GCD} & \multicolumn{3}{c}{PIM} \\ \cmidrule(l){2-4} \cmidrule(l){5-7}
     & All & Old & New & All & Old & New \\ \midrule
    w/o AGCD & 48.69 & 57.00 & 40.45 & 48.84 & 55.34 & 43.64 \\
    \texttt{Entropy} & 59.04 & 65.36 & 52.78 & 60.96 & 67.79 & 54.19 \\
    \texttt{BADGE} & 61.08 & 62.66 & 59.52 & 62.32 & \textbf{68.94} & 55.77 \\ \midrule
    \RC{30}Ours & \textbf{63.00} & \textbf{65.88} & \textbf{60.14} & \textbf{63.74} & 65.50 & \textbf{61.99} \\ \bottomrule
    \end{tabular}
    }
\end{table}

\paragraph{Comparison with more recent and open-set AL strategies.}
We adapt recent methods to AGCD, including  LfOSA~\cite{ning2022active}, MQ-Net~\cite{park2022meta}, ConAL~\cite{du2022contrastive}, and one additional standard AL method ALFA-Mix~\cite{parvaneh2022active}, results are shown in Table~\ref{tab:appendix-comparision-recent-al}. Because open-set AL merely cares about `Old ACC', and treats new classes as noise/outliers, it aims to detect/filter them and mainly query old classes. Instead, our approach further clusters new classes. As a result, open-set AL generally selects fewer samples from new classes, resulting in even worse performance than standard AL baselines in AGCD.

\begin{table}[!t]
    \setlength\tabcolsep{3pt}
    \centering
    \renewcommand{\arraystretch}{1.0}
    \caption{Comparison with \colorbox{color2}{open-set} \textcolor{blue}{and} \colorbox{color3}{standard} AL methods.}
    \vspace{-10pt}
    \label{tab:appendix-comparision-recent-al}
    \resizebox{.98\linewidth}{!}{
    \begin{tabular}{@{}cccccccc@{}}
    \toprule
    \multirow{2}{*}{Methods} & \multirow{2}{*}{Venue} & \multicolumn{3}{c}{CUB} & \multicolumn{3}{c}{SCars} \\ \cmidrule(l){3-5} \cmidrule(l){6-8}
     &  & All & Old & New & All & Old & New \\ \midrule
    \cellcolor{color2}\texttt{LfOSA} & CVPR'22 & 59.39 & 62.34 & 56.46 & 42.66 & 53.60 & 32.10 \\
    \cellcolor{color2}\texttt{ConAL} & T-PAMI'22 & 62.00 & 64.25 & 59.76 & 46.54 & \textbf{58.89} & 34.62 \\
    \cellcolor{color2}\texttt{MQ-Net} & NIPS'22 & 64.69 & \textbf{66.75} & 62.65 & 45.21 & 55.12 & 36.65 \\
    \cellcolor{color3}\texttt{ALFA-Mix} & CVPR'22 & 63.32 & 66.64 & 60.03 & 46.81 & 54.67 & 39.22 \\ \midrule
    \RC{30}Ours & This Work & \textbf{66.62} & 66.54 & \textbf{66.70} & \textbf{48.36} & 57.73 & \textbf{39.34} \\ \bottomrule
    \end{tabular}
    }
\end{table}


\begin{figure*}[!t]
    \centering
    \includegraphics[width=0.9\linewidth]{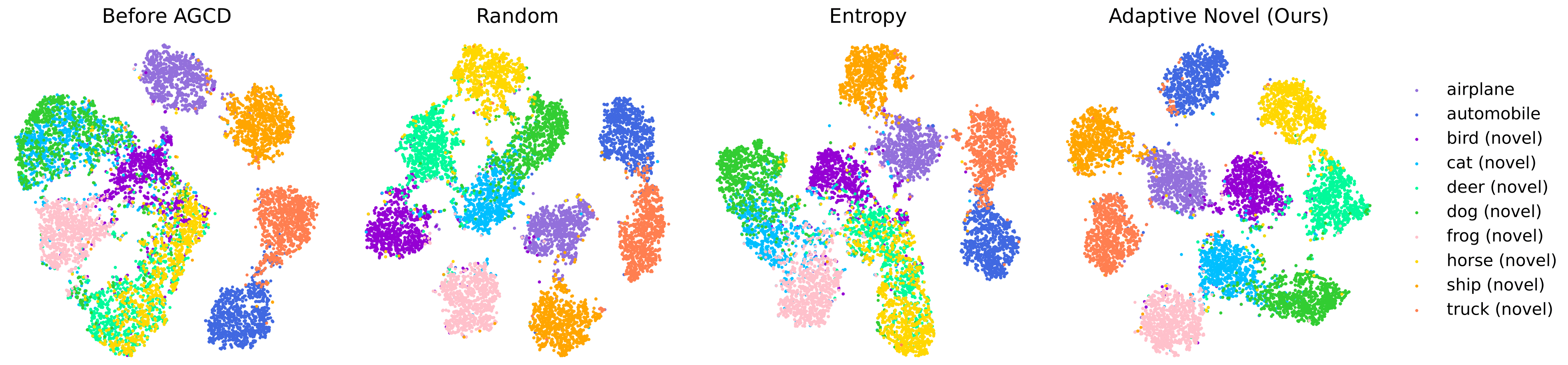}
    \caption{t-SNE~\cite{van2008visualizing} feature visualization of different strategies on CIFAR10.}
    \label{fig:appendix-vis-tsne-cifar10}
\end{figure*}

\begin{figure*}[!t]
    \centering
    \includegraphics[width=.9\linewidth]{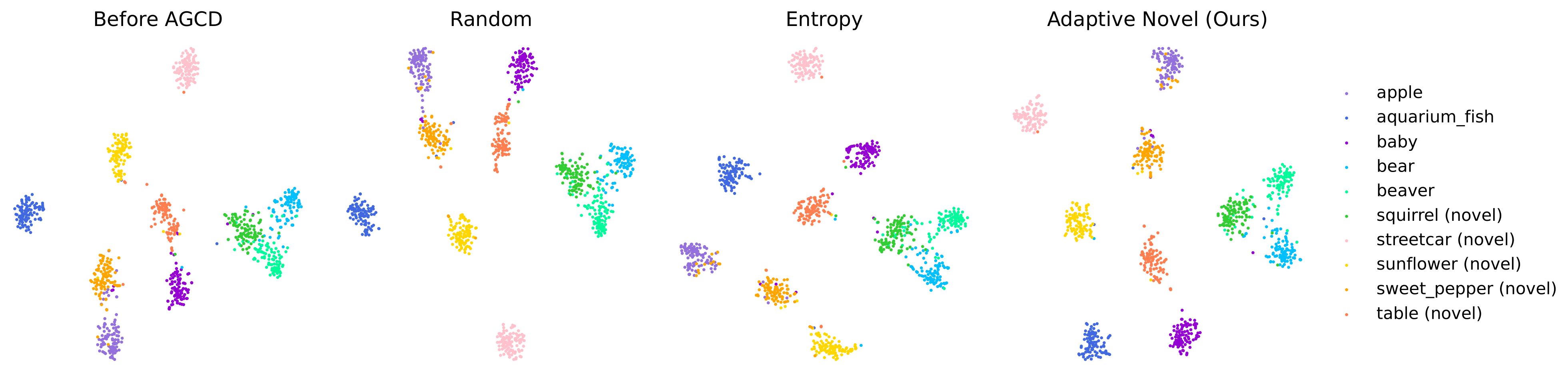}
    \caption{t-SNE~\cite{van2008visualizing} feature visualization of different strategies on CIFAR100.}
    \label{fig:appendix-vis-tsne-cifar100}
\end{figure*}

\begin{figure*}[!t]
    \centering
    \includegraphics[width=.9\linewidth]{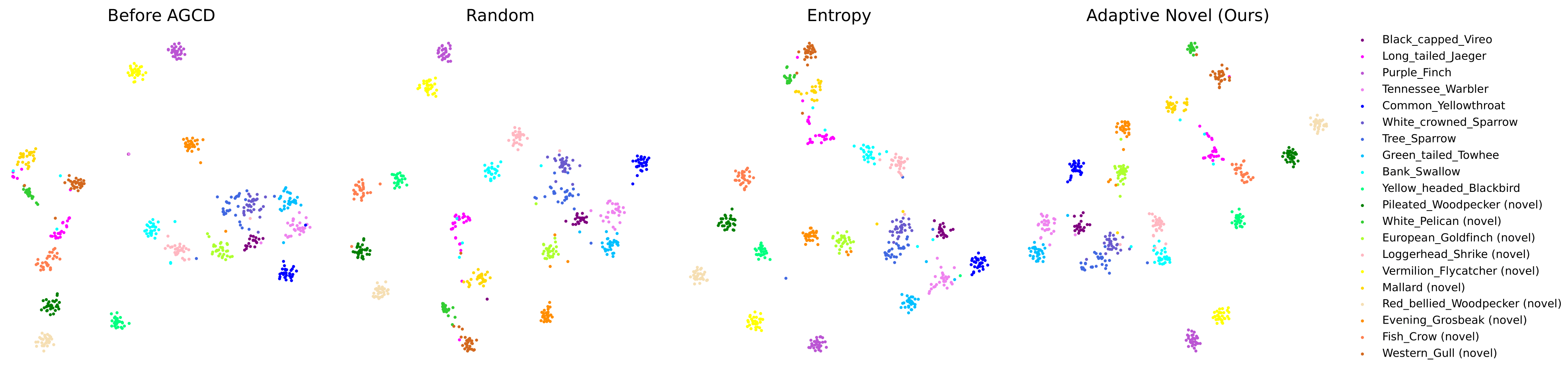}
    \caption{t-SNE~\cite{van2008visualizing} feature visualization of different strategies on CUB.}
    \label{fig:appendix-vis-tsne-cub}
\end{figure*}

\section{Visualization of Feature Spaces}
\label{sec:appendix-qualitative-results}

We visualize features of CIFAR10 using t-SNE~\cite{van2008visualizing} in Fig.~\ref{fig:appendix-vis-tsne-cifar10}. Original GCD (Left) suffers from severe confusion classes (\eg, ``deer" and ``horses''), while \texttt{Random} and \texttt{Entropy} struggle to select informative samples, resulting in overlapped cluster boundaries. Instead, our approach achieves clear inter-class separation. We further visualize the feature space on CIFAR100 and CUB using t-SNE~\cite{van2008visualizing}. For CIFAR100, we randomly selected 10 classes (5 old classes and 5 new classes), while for CUB, 20 classes in total (10 old classes and 10 new classes) for visualization, results are shown in Fig.~\ref{fig:appendix-vis-tsne-cifar100} and Fig.~\ref{fig:appendix-vis-tsne-cub}. As in Fig.~\ref{fig:appendix-vis-tsne-cifar100}, there are many confusing classes on which the model behaves ambiguously before AGCD, \eg, ``beaver'', ``squirrel'' and ``bear''. When the model is trained on the newly labeled data queried by our method, it could achieve relatively more separated clusters among the classes. Additionally, it is observable in Fig.~\ref{fig:appendix-vis-tsne-cifar100} and Fig.~\ref{fig:appendix-vis-tsne-cub} that our method could generally bring about more compact class-wise clusters on both datasets.

\end{document}